\documentclass[lettersize,journal]{IEEEtran}
\usepackage{amsmath,amsfonts}
\usepackage{algorithmic}
\usepackage{algorithm}
\usepackage{array}
\usepackage[caption=false,font=normalsize,labelfont=sf,textfont=sf]{subfig}
\usepackage{textcomp}
\usepackage{stfloats}
\usepackage{url}
\usepackage{verbatim}
\usepackage{graphicx}
\usepackage{cite}
\usepackage{xcolor}
\usepackage{booktabs}
\usepackage{multirow}
\usepackage{amsmath}
\usepackage{amssymb}
\usepackage{makecell}
\usepackage[colorlinks=true, linkcolor=black, urlcolor=magenta, citecolor=black]{hyperref}
\newcommand{\emailaddr}[1]{\nolinkurl{#1}}

\begin{document}

\title{S²GS: Structured Sparse Gaussian Streaming for Efficient Free-Viewpoint Video Reconstruction on Edge-IoT Devices}

\author{Yiwei Li, Jiannong Cao, \textit{Fellow, IEEE}, Weixun Gao, Rui Cao, Songye Zhu, Yinfeng Cao, and Mingjin Zhang
        
\thanks{%
Yiwei Li, Jiannong Cao, Weixun Gao, Rui Cao, Yinfeng Cao, and Mingjin Zhang
are with the Department of Computing, The Hong Kong Polytechnic University,
Hong Kong SAR, China
(e-mail:
\emailaddr{yi-wei.li@connect.polyu.hk};\allowbreak\space
\emailaddr{jiannong.cao@polyu.edu.hk};\allowbreak\space
\emailaddr{henry.gao@connect.polyu.hk};\allowbreak\space
\emailaddr{ruics.cao@connect.polyu.hk};\allowbreak\space
\emailaddr{csyfcao@comp.polyu.edu.hk};\allowbreak\space
\emailaddr{cs-mingjin.zhang@polyu.edu.hk}). 
Songye Zhu is with the Department of Civil and Environmental Engineering, The Hong Kong Polytechnic University,
Hong Kong SAR, China
(e-mail:
\emailaddr{songye.zhu@polyu.edu.hk}).
}

}

\markboth{Journal of \LaTeX\ Class Files,~Vol.~14, No.~8, August~2021}%
{Shell \MakeLowercase{\textit{et al.}}: A Sample Article Using IEEEtran.cls for IEEE Journals}


\maketitle

\begin{abstract}
Streaming reconstruction of Free-Viewpoint Videos (FVVs) supports immersive Internet of Things (IoT) services, such as telepresence and digital twin visualization. Existing methods suffer from high per-frame optimization time and large storage footprints, limiting deployment on resource-constrained Edge-IoT devices. To address these challenges, we propose Structured Sparse Gaussian Streaming (S²GS), an FVV reconstruction framework that exploits structure-aware temporal sparsity to selectively update Gaussian residuals, enabling efficient streaming without compromising visual fidelity. In the spatial domain, a streaming octree hierarchically organizes Gaussian residuals, capturing spatial correlations that guide residual updates. In the temporal domain, a structured gating mechanism, comprising hierarchical feature propagation (HFP) and Gumbel-Sigmoid sampling, converts hierarchical dynamic cues into sparse residual update decisions under differentiable optimization. A multi-level discrete scheme is further adopted to provide fine-grained control over residual updates while preserving intricate dynamic details. Extensive experiments across consumer GPUs, industrial edge IoT devices, and a physical telepresence testbed demonstrate that S$^2$GS consistently reduces per-frame optimization time and storage footprint while maintaining competitive visual quality. Compared with QUEEN, S$^2$GS reduces per-frame optimization time by 59\% and storage costs by 85\% on an RTX 4090 GPU. On the Jetson AGX Orin, S²GS delivers the highest rendering throughput (60+ FPS) and the lowest energy consumption among the evaluated methods, demonstrating its potential for deployment in resource-constrained systems. Code: \href{https://github.com/liyw420/S2GS}{this URL}
\end{abstract}


\section{Introduction}
\label{sec:introduction}

Immersive Internet of Things (IoT) services, such as telepresence \cite{li2022internet, huang2024enhancing} and digital twin (DT) visualization \cite{fu2026generative}, require realistic representations that capture the dynamic evolution of physical environments. Free-viewpoint video (FVV) provides this foundation by reconstructing dynamic 3D scenes that users can interactively explore from freely selected viewpoints \cite{zhou2026gausshead}. Streaming FVV reconstruction, in which scenes are updated frame by frame, further enables online responses and reduced per-frame packet sizes compared to offline FVV approaches, making it promising for efficient operation and transmission in bandwidth-constrained edge IoT systems \cite{wei2023vsas, wei2024fewvv, tu2026civs}.

\begin{figure}[t]
  \centering
   \includegraphics[width=1.0\linewidth]{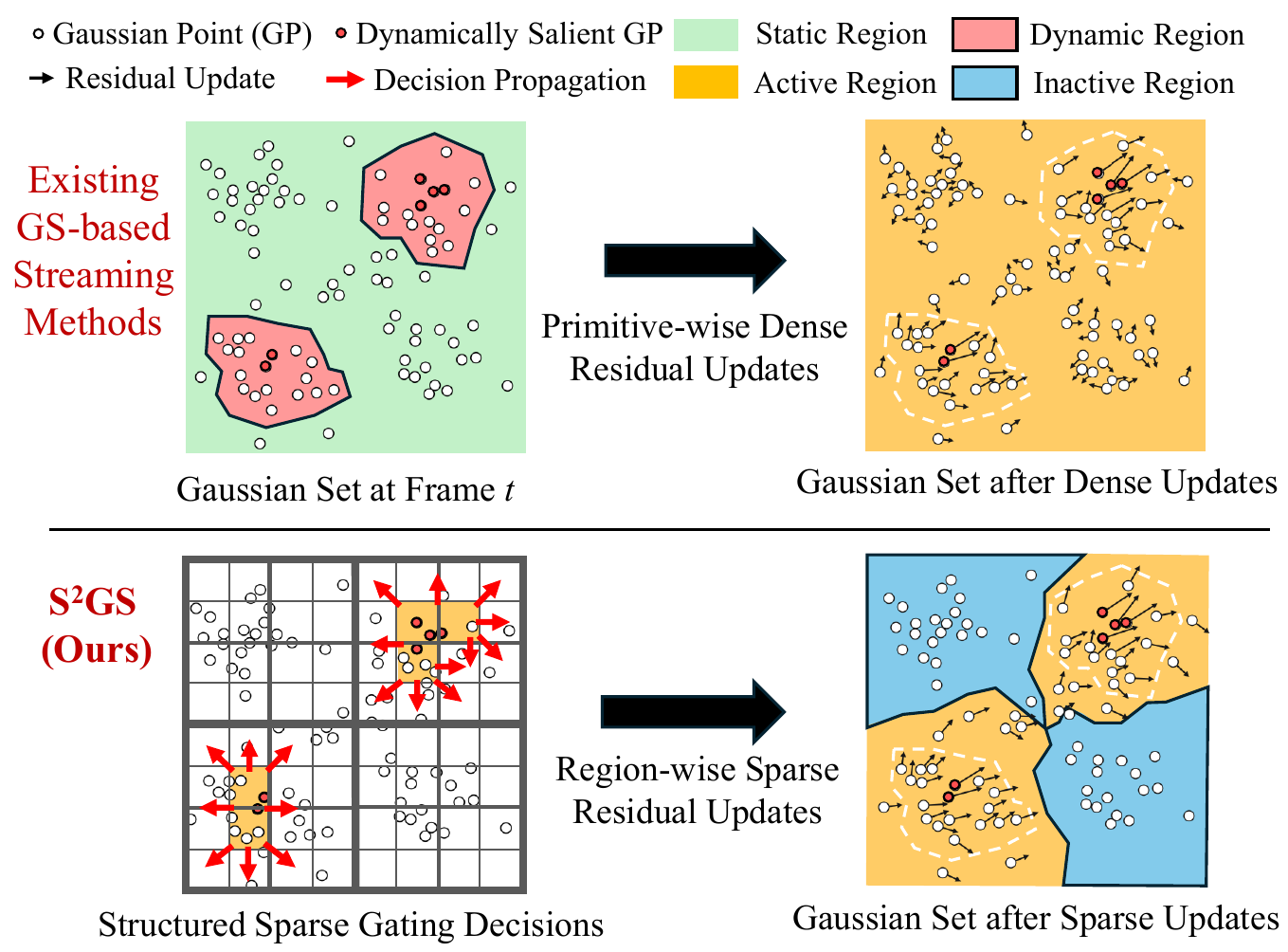}
    \caption{
    Schematic comparison of Gaussian residual update paradigms.
    \textbf{Top:} Existing GS-based streaming methods perform primitive-wise dense residual updates. \textbf{Bottom:} S$^2$GS exploits spatially correlated temporal sparsity to select a small number of hierarchical regions and perform region-wise sparse residual updates only within active regions.
    }
   \label{concept}
\end{figure}

\begin{figure*}[t]
  \centering
   \includegraphics[width=1.0\linewidth]{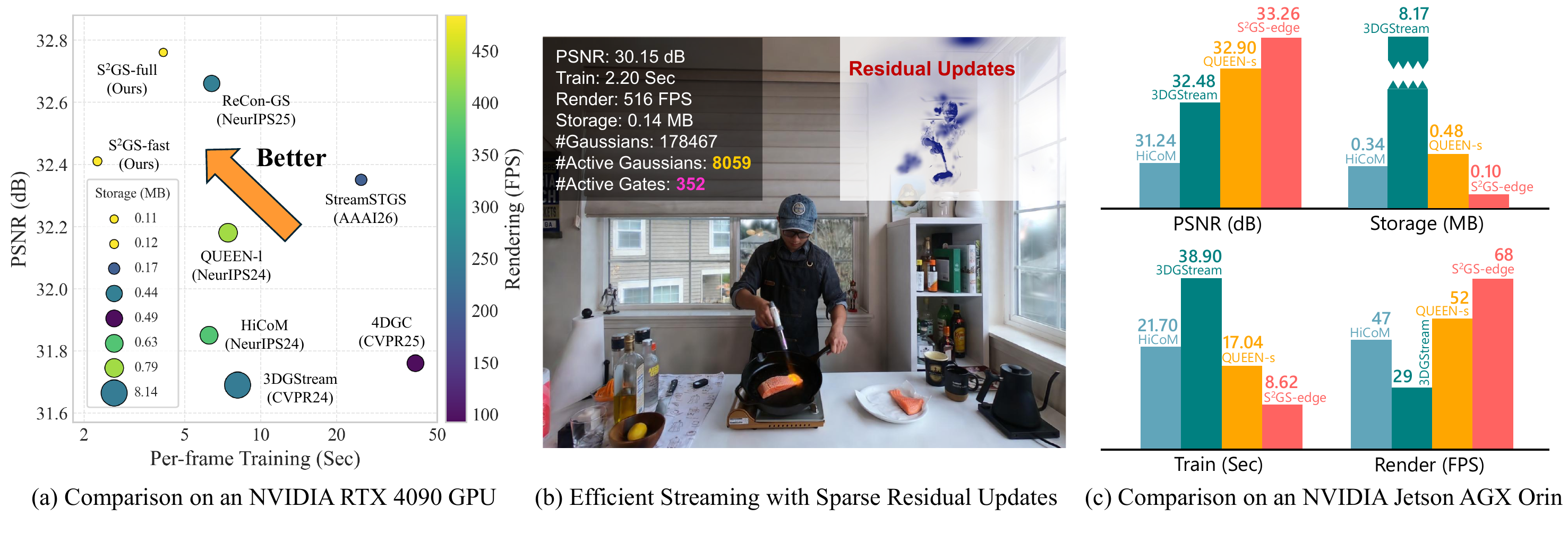}

    \caption{Overview and performance of the proposed structured sparse Gaussian streaming framework (S²GS). S²GS delivers high-quality FVV reconstruction with improved efficiency in per-frame training, rendering, and storage. (a) Quantitative comparison on the N3DV dataset, where PSNR, per-frame training time, storage cost (circle size), and rendering throughput (color) are jointly compared. (b) Visualization of sparse residual updates on the challenging Flame Salmon scene. S$^2$GS-fast concentrates updates on dynamic regions, using only 4.52\% active Gaussians and 4.37\% active gates for efficient FVV streaming. (c) Edge deployment on the N3DV dataset at 1/8 resolution, where S²GS-edge outperforms baseline methods in visual quality, storage, per-frame training time, and rendering throughput.}
   \label{teaser}
\end{figure*}

Recent advances in Gaussian Splatting (GS) have significantly improved the efficiency of FVV reconstruction over neural volumetric methods such as NeRF \cite{mildenhall2021nerf} and its variants \cite{li2022streaming, li2022neural, song2023nerfplayer}. By leveraging an explicit point-based representation, GS enables photorealistic visualization with efficient training and rendering. Nevertheless, existing GS-based streaming approaches still suffer from high per-frame optimization time and large storage footprints, which limit their deployment on resource-constrained edge-IoT devices. As shown in Fig.~\ref{teaser}(c), when deployed on the NVIDIA Jetson AGX Orin, current methods cannot simultaneously maintain high visual quality, low optimization time, and compact storage. For example, QUEEN~\cite{girish2024queen}, one of the most recent approaches, requires more than 17 seconds of per-frame optimization time (reported as Train (Sec)) and 0.48 MB of storage per frame. At 30 FPS, this amounts to 14.4 MB of residual data per second, introducing accumulated storage overhead for edge-assisted telepresence systems or industrial DTs~\cite{hu2025varfvv, liao2025graph}.

We argue that the trade-off among quality, optimization time, and storage partly arises from the dense residual updates in existing GS-based methods (Fig. \ref{concept}, top). Most prior approaches do not explicitly identify which Gaussians require residual updates across frames; instead, they improve compactness or compression efficiency by reorganizing Gaussian primitives~\cite{sun20243dgstream, gao2024hicom, chen2025motion} or by quantizing Gaussian attributes and encoding residuals~\cite{li2025gifstream, girish2024queen, hu20254dgc, zheng4dgcpro}. Although effective, these methods primarily reduce the cost of representing or transmitting residuals, rather than determining where residual updates are necessary. They overlook a key property: \textbf{\textit{the temporal sparsity of Gaussian residual updates}}, namely, that only a small subset of Gaussians undergo substantial residual updates between consecutive frames, while most primitives remain static or change slightly. As illustrated in Fig.~\ref{teaser}(b), only 4.52\% of Gaussian primitives are active for residual updates in a representative FVV frame (see cross-dataset statistics in Appendix \ref{Definition and Statistical Analysis of Structured Sparsity}). Under dense update formulations, computation and storage are allocated to Gaussians with little contribution to scene dynamics, resulting in redundant optimization and storage overhead. This observation suggests that compact representation and residual compression alone are insufficient; explicit modeling of sparse Gaussian residual updates provides a complementary direction for efficient FVV streaming.

A further observation is that the temporal sparsity of Gaussian residual updates is not necessarily modeled independently across Gaussian primitives; instead, it often exhibits spatially correlated patterns. As illustrated in Fig.~\ref{teaser}(b), the active Gaussians can be summarized by only 4.37\% active gates (see cross-dataset statistics in Appendix \ref{Definition and Statistical Analysis of Structured Sparsity}), where residual updates are concentrated in local regions rather than uniformly scattered over all primitives. This region-level sparsity suggests that residual updates can be modeled more efficiently beyond primitive-wise updates. In particular, a hierarchical spatial organization provides a natural structure for representing and propagating local sparse patterns from fine primitives to coarser regions  \cite{wei2025octgpt}. Together with the first observation that only a small subset of Gaussians requires substantial residual updates, this finding suggests FVV streaming to move beyond dense primitive-wise residual updates and exploit scene structure to guide sparse residual updates. This motivates our objective: \textbf{\textit{developing a structure-aware sparsity framework for FVV streaming that induces sparse Gaussian residual updates through the spatial hierarchy of the scene, thereby improving efficiency for deployment on edge-IoT devices.}}

To achieve this goal, we propose the Structured Sparse Gaussian Streaming (S$^2$GS) framework. As shown in Fig. \ref{concept} bottom, the core idea is to model temporally sparse residual updates over a fixed hierarchical structure, so that sparse update patterns follow the region-wise organization of Gaussian primitives. In the spatial domain, the streaming octree hierarchically organizes Gaussian residuals and captures spatial correlations among dynamic variations. In the temporal domain, rather than learning dense primitive-wise residual updates, we formulate residual learning as a structure-aware sparse update process, in which region-wise residual updates are selectively assigned to dynamic Gaussians.

Nevertheless, realizing this formulation is nontrivial. The first challenge is how to propagate dynamic confidences, which indicate whether a Gaussian primitive requires residual updates, across the spatial hierarchy while preserving fine-grained temporal changes. The second challenge is how to convert these structure-aware confidences into sparse residual update decisions under differentiable optimization. To address these challenges, we develop a structured gating mechanism with two coupled components. First, hierarchical feature propagation (HFP) aggregates dynamic cues along the spatial hierarchy, producing structure-aware dynamic confidences that capture both local temporal variations and the spatial consistency of neighboring dynamic regions. Second, Gumbel-Sigmoid sampling converts these confidences into continuous soft gates, which relax discrete residual update decisions and enable differentiable optimization. Since strict binary discretization is overly aggressive and suppresses regions that require residual updates, we further introduce a multi-level discrete gating scheme that replaces hard 0/1 gate decisions. This design provides finer-grained control over the strength of residual updates and helps preserve intricate dynamic details. In this way, sparsity is encouraged not only by a regularizer, but also by structured-guided residual update decisions.

We evaluate S²GS against state-of-the-art (SOTA) baselines on a consumer-grade GPU, an industrial-grade edge-IoT device, and a real-world physical testbed for telepresence. The main contributions of this work are summarized as follows.

\begin{enumerate}
    \item We identify structured sparsity in Gaussian residual updates as a key property for efficient FVV streaming, showing that sparse residual updates are spatially correlated and can be guided by the scene hierarchy.
    \item We propose S$^2$GS, an FVV streaming framework that exploits structure-aware temporal sparsity to selectively update Gaussian residuals, enabling efficient streaming reconstruction without compromising visual fidelity.
    \item We develop a structured gating mechanism with two key components. HFP aggregates primitive-wise dynamic cues into region-wise confidences and propagates gate decisions through the hierarchy, while the multi-level discrete scheme mitigates gate collapse and maintains stable sparse updates with intricate dynamic details.
    \item Extensive experiments on challenging FVV datasets demonstrate that S$^2$GS maintains competitive visual quality while reducing the storage footprint and improving training and rendering efficiency. Additional evaluations on an edge device and a physical testbed further validate its low resource costs and practical potential in resource-constrained edge-IoT systems.
\end{enumerate}

\section{Related Work}

\subsection{Offline FVV Reconstruction}
The recent emergence of Gaussian Splatting~\cite{kerbl20233d} has significantly advanced offline FVV reconstruction due to its efficiency and flexibility. For example, STG~\cite{li2024spacetime} and Ex4DGS~\cite{lee2024fully} model dynamic motions of Gaussian primitives using polynomial functions. Several representative methods, including E-D3DGS~\cite{bae2024per} and 4DGaussians~\cite{wu20244d}, decouple dynamic scenes into a canonical space and a deformation field. 4DGS~\cite{yang2023real} further extends Gaussian Splatting to higher dimensions by introducing spatiotemporally coherent 4D Gaussians. Meanwhile, Grid4D~\cite{xu2024grid4d} leverages hash encoding together with an attention mechanism to capture object deformations. Although these approaches achieve impressive visual quality and real-time rendering, they typically require access to complete video sequences for model optimization, which fundamentally limits their applicability to live FVV streaming.

\subsection{Online FVV Reconstruction}

Online FVV reconstruction methods sequentially optimize scenes over time, enabling live-streaming input processing and incremental model updates. StreamRF~\cite{li2022streaming} and NeRFPlayer~\cite{song2023nerfplayer} extend implicit neural representations to support streamable rendering. Similarly, 3DGStream~\cite{sun20243dgstream} adapts 3D Gaussian Splatting for FVV streaming by incorporating multi-resolution hash encoding. HiCoM~\cite{gao2024hicom} and ReCon-GS~\cite{furecon} introduce explicit motion fields to improve model efficiency. In contrast, QUEEN~\cite{girish2024queen} and 4DGC~\cite{hu20254dgc} employ differentiable quantization to compress Gaussian residual attributes. IGS~\cite{yan2025instant} further leverages a pretrained motion network to enable generalizable FVV streaming, while ComGS~\cite{chen2025motion} exploits motion locality to reduce memory overhead. Despite their impressive results, these methods largely overlook the inherent structured sparsity for residual updates in FVVs. Additionally, most existing approaches optimize only specific aspects (e.g., quality, model size), making them less suitable for deployment on resource-constrained edge-IoT devices, where comprehensive improvements in per-frame optimization, storage, and resource consumption are required for real-world IoT applications.

\subsection{Hierarchical Scene Representation}
Hierarchical representations have been widely explored to improve the scalability and rendering efficiency of Gaussian Splatting. For instance, Scaffold-GS~\cite{lu2024scaffold} organizes local Gaussians with anchors, while Octree-GS~\cite{renoctree} extends this idea for level-of-detail (LOD) rendering. LODGE~\cite{kulhaneklodge} further introduces a hierarchical LOD representation for efficient large-scale rendering. In contrast to prior methods that use hierarchical structures to improve scene representation and rendering, S$^2$GS employs a fixed hierarchy to organize and selectively update Gaussian residuals, while propagating sparse gating decisions across the hierarchy during FVV streaming.

\begin{figure*}[t]
  \centering
   \includegraphics[width=1.0\linewidth]{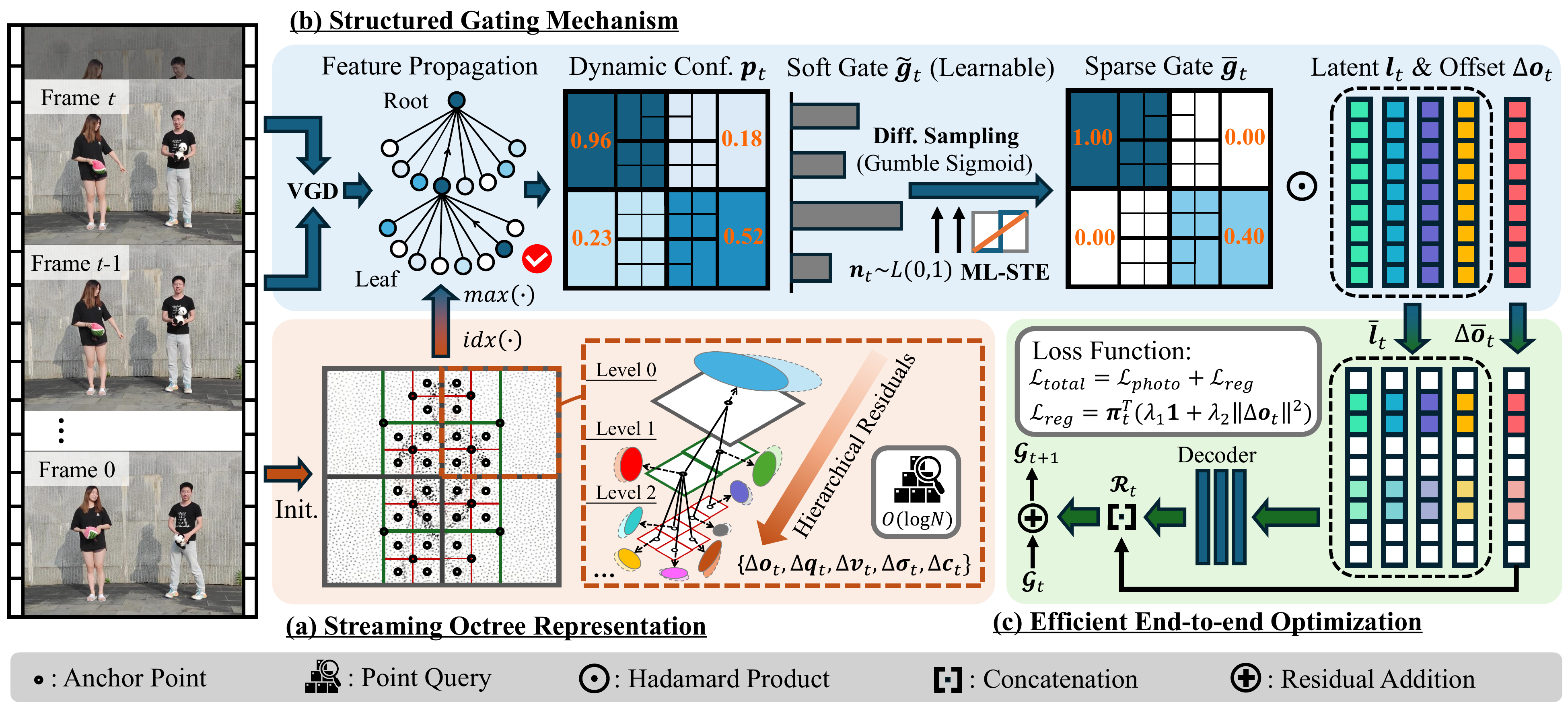}

   \caption{Overview of the proposed S²GS framework. (a) The streaming octree representation is initialized using root Gaussian primitives from the first frame and subsequently represents FVVs with multi-resolution grids, enabling hierarchical allocation of Gaussian residuals and efficient point queries for FVV streaming. (b) A structured gating mechanism is introduced to induce sparse Gaussian residual updates through hierarchical feature propagation, differentiable sampling with Gumbel-Sigmoid, and multi-level (ML) discretization using a straight-through estimator (STE). (c) A sparse regularization loss is incorporated to enable efficient end-to-end optimization of structured gates and sparse residual updates.}
   \label{pipeline}
\end{figure*}

\section{Proposed Method}

\subsection{Problem Formulation}
Given a set of synchronized multi-view image sequences $\{{\mathcal{I}^{T-1}_{t=0}}\}$ with known camera poses, where $t$ indexes $T$ frames, our objective is to sequentially synthesize photorealistic FVVs in a streaming manner at each time step. Let $\mathcal{G}_{t}$ denote the set of 3D Gaussians at time $t$, defined as $\mathcal{G}_{t} = \{\boldsymbol{\mu}_t, \boldsymbol{q}_t, \boldsymbol{s}_t, \boldsymbol{\sigma}_t, \boldsymbol{c}_t\}$, where $\boldsymbol{\mu}_t \in \mathbb{R}^{N\times3}$, $\boldsymbol{q}_t \in \mathbb{R}^{N\times4}$, $\boldsymbol{s}_t \in \mathbb{R}^{N\times3}$, $\boldsymbol{\sigma}_t \in \mathbb{R}^{N}$, and $\boldsymbol{c}_t$ represent the position vector, rotation quaternion, scaling factor, opacity, and color coefficient of the Gaussian set with $N$ primitives, respectively \cite{kerbl20233d}. The Gaussian set $\mathcal{G}_{t+1}$ for the next frame is then obtained by learning residuals $\mathcal{R}_{t}$ with respect to the optimized $\mathcal{G}_{t}$ from the previous frame:
\begin{align}
{{\mathcal{G}_{t+1}} = {\mathcal{G}_{t} + \mathcal{R}_{t}}}, \label{eq4}
\end{align}
where $\mathcal{R}_{t} = \{\Delta\boldsymbol{\mu}_t ,\Delta\boldsymbol{q}_t , \Delta\boldsymbol{s}_t , \Delta\boldsymbol{\sigma}_t, \Delta\boldsymbol{c}_t\}$ consists of residuals for each Gaussian attribute. The Gaussian set $\mathcal{G}_{0}$ at the first frame is initialized and optimized from point clouds generated by 3D reconstruction pipelines (e.g., COLMAP \cite{schoenberger2016sfm, schoenberger2016mvs}). This work aims to develop a structure-aware sparsity framework that guides sparse Gaussian residual updates by the underlying spatial hierarchy of the scene, thereby improving efficiency for practical deployment on resource-constrained edge-IoT infrastructures.

\subsection{Overview}
To enable FVV streaming on resource-limited edge-IoT devices, we propose S²GS, a GS‑based streaming framework that leverages end‑to‑end learnable structured sparsity to achieve fast and compact reconstruction. The overview of S²GS is illustrated in Fig. \ref{pipeline}, which consists of three components.

\subsubsection{Streaming Octree Representation}
This forms the basic spatial structure of the FVV streaming pipeline. By leveraging multi‑resolution grids, it hierarchically organizes Gaussian residuals to provide spatial correlations among dynamic variations. The structure further supports efficient streaming queries for HFP during optimization.

\subsubsection{Structured Gating Mechanism}
The core component of S²GS that formulates the residual learning as a structure-aware sparse update process. It first estimates per‑point dynamic confidences from consecutive frames and propagates them through the spatial hierarchy. Probabilistic gating decisions are then realized via Gumbel-Sigmoid relaxation paired with a multi-level straight‑through estimator (STE) \cite{bengio2013estimating}, thereby yielding discrete and learnable gates.

\subsubsection{Efficient End-to-End Optimization}
This component enables end‑to‑end optimization of both structured gates and attributes of Gaussian residuals. A regularization loss is applied to jointly optimize gate decisions and residual updates.

\subsection{Streaming Octree Representation}

Inspired by the octree-based hierarchical decomposition \cite{renoctree,wei2025octgpt}, we introduce a streaming octree representation. The framework is initialized with \textit{Root Gaussian Primitives}, denoted as $\mathcal{G}_{\mathcal{R}}$, which is a static set of 3D Gaussians reconstructed from the first frame of multi-view inputs.

\subsubsection{Hierarchical Gaussian Residual Allocation} 

Given the global spatial bounds of $\mathcal{G}_{\mathcal{R}}$, an octree is first constructed within the bounded volume, and each anchor is initialized at the centroid of its corresponding voxel~\cite{renoctree}. Based on this spatial structure, we represent FVV streaming using fixed anchors with temporally varying and hierarchical Gaussian residuals. The residuals of the Gaussian set at frame $t$ are defined as:
\begin{align}
    \mathcal{R}_t &= \bigl\{\, \Delta\boldsymbol{o}_t,\; \Delta\boldsymbol{q}_t,\; \Delta\boldsymbol{v}_t,\; \Delta\boldsymbol{\sigma}_t,\; \Delta\boldsymbol{c}_t \,\bigr\}, \label{eq:residual_set} \\
    \Delta\boldsymbol{v}_t &= \operatorname{concat}\bigl( \Delta\boldsymbol{s}_t,\; \Delta\boldsymbol{f}_t \bigr). \label{eq:v_def}
\end{align}
where $\operatorname{concat}(\cdot)$ denotes the concatenation operation, $\Delta\boldsymbol{o}_{t} \in \mathbb{R}^{N \times 3}$ represents the offset vector of the Gaussian residuals, and $\Delta\boldsymbol{f}_t \in \mathbb{R}^{N \times 3}$ is the updated anchor-dependent scaling factor. As illustrated in Fig.\ref{pipeline} (a), the proposed streaming octree representation hierarchically organizes Gaussian residuals and captures spatial correlations across different levels of dynamic variation. This allocation establishes the structural foundation for the subsequent structure-aware sparse residual updates.

\subsubsection{Efficient Streaming Queries}
Unlike prior works \cite{sun20243dgstream, gao2024hicom, girish2024queen, furecon} that perform online pruning and densification of their spatial data structures, S$^2$GS keeps the spatial hierarchy fixed once established. The hierarchy remains unchanged throughout the training of subsequent frames. This fixed-hierarchy design allows all per-frame Gaussian residuals to be associated with persistent, pre-allocated anchors. Consequently, the streaming octree lookup operation in S²GS is ${O}(\operatorname{log}N)$, where $N$ is the number of Gaussian primitives. This logarithmic complexity enables efficient point queries within the spatial hierarchy, making S²GS a practical solution for spatial indexing in the subsequent FVV streaming process.

\subsection{Structured Gating Mechanism}\label{Sec. D}
Establishing an effective gating mechanism requires propagating dynamic confidences and optimizing gate decisions in a differentiable manner. In addition, gate learning in S$^2$GS accounts for structured sparsity, allowing propagation across the spatial hierarchy while avoiding unnecessary residual updates in static regions. For notational simplicity, we omit the subscript $t$ in this section, as all subsequent calculations are performed with respect to frame $t$.

\subsubsection{Dynamic Confidence Estimation} 
We estimate the confidence of Gaussian primitives being dynamic based on the viewspace gradient difference (VGD)~\cite{girish2024queen, chen2025motion}. Specifically, during FVV streaming, we first compute the VGD $\boldsymbol{m} \in \mathbb{R}^{N \times 2}$ from two consecutive Gaussian sets, each containing $N$ primitives. The dynamic confidence of the $i$-th Gaussian is computed as follows:
\begin{align}
    p_i = \sigma\Biggl(
        \frac{ |\boldsymbol{m}_i| - \operatorname{median}(|\boldsymbol{m}|) }
             { \operatorname{MAD}(|\boldsymbol{m}|) + \varepsilon }
    \Biggr),
    \label{eq:gate_prob}
\end{align}
where $\sigma(\cdot)$ denotes the sigmoid function, $\operatorname{MAD}(\cdot)$ denotes the median absolute deviation, and $\varepsilon$ is a small positive constant for numerical stability, set to $2 \times 10^{-4}$.

\subsubsection{Hierarchical Feature Propagation} 
Once $p_i$ is estimated, the cached features are further aggregated for efficient gate learning. A straightforward strategy is to discard voxels with confidence values below a fixed threshold \cite{bai2023dynamic}. However, such threshold-based pruning may remove informative features and weaken spatiotemporal consistency, which is critical for reconstructing coherent dynamic scenes in FVV streaming. 

To address this issue, we propose a hierarchical feature propagation (HFP) module. Specifically, leaf-level confidences are propagated through the octree hierarchy and aggregated into root-level voxels by max pooling:
\begin{align}
    p_j = \max_{\{i \mid \operatorname{idx}(i) = j\}} p_i, \:\:  i \in \{0, 1, \dots, N-1\},
    \label{eqHFP}
\end{align}
where $j \in \{0, 1, \dots, R-1\}$ indexes the root-level voxels, and $\operatorname{idx}(\cdot)$ maps each Gaussian primitive $i$ to its corresponding root-level parent $j$. The aggregation strategy in Eq. (\ref{eqHFP}) is central to HFP. Unlike threshold-based pruning, which irreversibly removes low-confidence voxels before optimization, the proposed max-pooling scheme preserves salient dynamic cues within each local subtree and retains candidate motion evidence for subsequent gate learning. As a result, the root-level confidence provides a conservative indication of whether informative dynamics exist within a structured local region.

In addition, after gate sampling in Sec. \ref{section_dgs}, root-level gate decisions are distributed to all associated leaf-level Gaussians through the same index mapping. With HFP, the confidence of each root-level region is computed by aggregating the confidences of its associated leaf-level Gaussians. This allows gate learning to be performed at the region level while retaining fine-grained dynamic cues from leaf-level Gaussians. As a result, HFP preserves spatiotemporal consistency and reduces the computational cost of gate learning.

\subsubsection{Differentiable Gate Sampling}\label{section_dgs}
The gate assigned to each Gaussian residual update is a binary variable. The main challenge for differentiable gate sampling lies in the non-differentiability of discrete operators, such as $\operatorname{arg\, max}(\cdot)$ and $\operatorname{onehot}(\cdot)$~\cite{fang2024maskllm}. To address this issue, we leverage Gumbel Sigmoid \cite{maddison2017concrete}, a binary concrete distribution, to provide a continuous, differentiable relaxation of the discrete sampling process, enabling gradient‑based optimization of the gating mechanism:
\begin{align}
    {\tilde{g}_j} &= \sigma\Bigl( \frac{\log\bigl(p_j / (1 - p_j)\bigr) + n}{\tau} \Bigr), \label{eq:gumbel_sigmoid} \\
    n &\sim \mathrm{Logistic}(0, 1), \label{eq:logistic_noise}
\end{align}
where ${\tilde{g}_j}$ denotes the continuous soft gate, $n$ is a noise variable sampled from a Logistic distribution, and the temperature $\tau$ controls the hardness of the sampled gate and is set to 0.3.

While this formulation yields differentiable gates, it does not inherently induce sparsity. To obtain sparse, discrete gates during training while preserving gradient flow, we employ a straight‑through estimator (STE) \cite{bengio2013estimating}. Specifically, in the forward pass, the soft gate ${\tilde{g}_j}$ is rounded to generate a discrete, gradient‑stopped gate ${\bar{g}_j}$:
\begin{align}
{{\bar{g}_j}} = \operatorname{stop\_grad}(\operatorname{round}({\tilde{g}_j})),\label{eq_ste}
\end{align}
where $\operatorname{stop\_grad}(\cdot)$ denotes the gradient-stopping operation. In the backward pass, gradients are propagated as if the gate remained continuous ${\tilde{g}_j}$. 

\begin{figure}[t]
  \centering
   \includegraphics[width=1.0\linewidth]{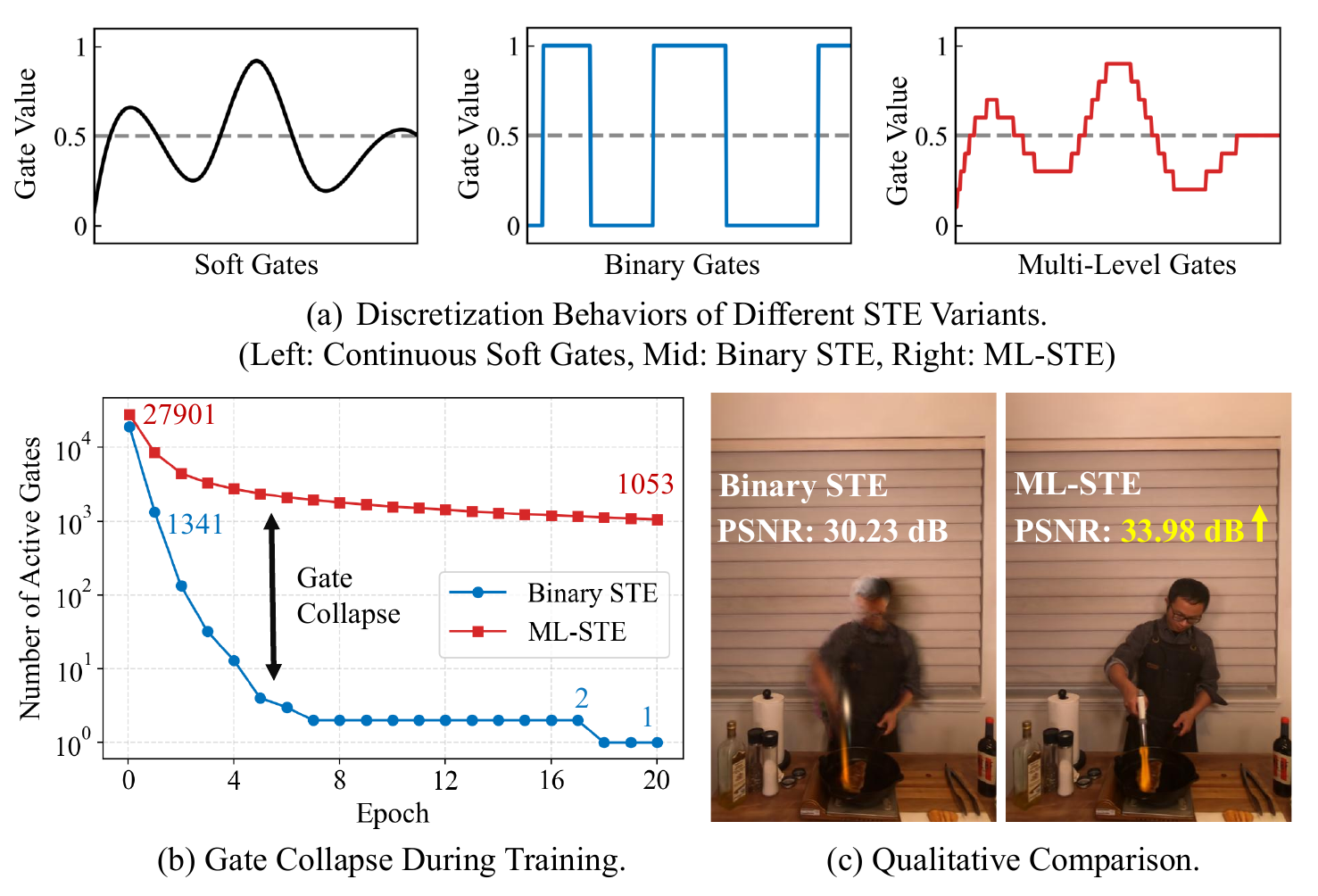}
   \caption{Analysis of different STE variants. (a) Continuous soft gates and their discretized counterparts produced by Binary STE and the proposed Multi-Level (ML) STE. Unlike Binary STE, ML-STE preserves finer-grained gate states after discretization. (b) Number of active gates over training epochs for a representative FVV frame. Binary STE rapidly leads to gate collapse, whereas ML-STE maintains stable sparse residual updates throughout training. (c) Visual comparison on the same FVV frame as in (b), where ML-STE achieves better reconstruction quality than Binary STE, particularly in the dynamic region.}
   \label{ste}
\end{figure}

In our experiments, however, we observed that directly rounding STE outputs to strict binary values of 0 or 1, as in Binary STE, can lead to gate collapse during training. Most gates rapidly become zero, preventing the model from assigning residual updates to regions with fine-grained temporal variations and severely degrading reconstruction quality, as illustrated in Fig.~\ref{ste}. To address this issue, we employ a multi-level (ML) STE scheme. It discretizes $\tilde{g}_j$ to one decimal place rather than enforcing strict 0/1 decisions. The resulting multi-level discrete gates retain the exact-zero state for sparse selection while providing control over the magnitude of active residual updates. This finer discretization mitigates gate collapse, maintains a stable sparse set of active gates throughout training, and preserves the residual capacity required to model intricate temporal dynamics, thereby improving visual fidelity.

Overall, this design enables the underlying gate confidence $p_j$ to be optimized end-to-end while preserving sparse residual update decisions at runtime, thereby ensuring stable and consistent gate behavior from training to rendering.

\subsection{Efficient End-to-End Optimization}
To encourage sparsity and regularize the magnitude of residual updates, we propose a joint regularization loss $\mathcal{L}_{\mathrm{reg}}$. It combines an approximate ${L_0}$ penalty that promotes exact zero residuals with an ${L_2}$ term that constrains the magnitude of non-zero offset updates, which is formulated as:
\begin{align}
\mathcal{L}_{\mathrm{reg}} = \sum_{i=0}^{N-1}{\pi_i} \cdot \Bigl( \lambda_{{1}} + \lambda_{{2}} \, \bigl\lVert  \Delta{\boldsymbol{o}}_i \bigr\rVert_2^2 \Bigr),\label{eq_sparse_loss}
\end{align}
where ${\pi_i} = 1- F_{p_i}(\eta)$ denotes the dynamic probability of Gaussian primitive $i$, $F_{p_i}(\eta)$ is the cumulative distribution function of ${p_i}$ evaluated at a threshold $\eta$, $\Delta{\boldsymbol{o}}_i$ is the offset update without sparsification, and $\lambda_{{1}}$, $\lambda_{{2}}$ are weighting coefficients. This formulation enables the model to jointly learn sparse and compact residual updates for Gaussian attributes while maintaining differentiability throughout optimization. 

Following \cite{girish2024queen}, residuals of Gaussian attributes are learned from latent vectors $\{\boldsymbol{l}_a| a\in\{\boldsymbol{q} , \boldsymbol{s} , \boldsymbol{\sigma}, \boldsymbol{c}\}\}$ by passing through compact linear decoders. The loss function is formulated as:
\begin{align}
\mathcal{L}_{\mathrm{total}} = \mathcal{L}_{\mathrm{photo}} + \mathcal{L}_{\mathrm{reg}},
\end{align}
where the photometric loss $\mathcal{L}_{\mathrm{photo}}$ is computed by comparing rendered images across time steps with ground-truth images using a combination of ${L}_{1}$ loss and D-SSIM loss \cite{kerbl20233d}.

\section{Experiments and Results}
\subsection{Experimental Goal}

We conduct experiments to answer the following questions.

\begin{enumerate}
    \item \textit{RQ-1 (Quality):} Does S²GS achieve competitive visual quality compared to state-of-the-art methods?
    \item \textit{RQ-2 (Efficiency):} Does S²GS outperform existing baseline methods in model complexity, storage size, training efficiency, and rendering throughput?
    \item \textit{RQ-3 (Ablation Study):} Does each proposed component of S²GS contribute positively to its overall results?
    \item \textit{RQ-4 (Deployability):} Is S²GS easily deployable on edge-IoT devices while meeting the requirements for quality, timeliness, and resource efficiency?
    \item \textit{RQ-5 (Case Study):} How does S²GS perform when applied in real-world edge-IoT environments?
\end{enumerate}

\begin{table*}[!tp]
\centering
\caption{Quantitative comparisons on the N3DV dataset at 1/2 resolution. For all methods, per-frame metrics are averaged across all frames. Results for methods marked with $^{\dagger}$ are taken from the original papers. All experiments are conducted on an NVIDIA RTX 4090 GPU. \textcolor{red}{\textbf{Red}} and \textcolor{blue}{\underline{blue}} denote the best and second-best results, respectively, within each category.}
\fontsize{7.8}{10}\selectfont
\setlength{\tabcolsep}{2.0pt}
\renewcommand{\arraystretch}{1.05}
\begin{tabular}{
l|  
l|  
ccccccc|  
c|  
c|  
c|  
c   
}
\toprule
& \multirow{2}{*}{Methods}
& \multicolumn{7}{c|}{PSNR (dB)$\uparrow$ / SSIM$\uparrow$}
& \multirow{2}{*}{\makecell{\#Gaussians \\(k)$\downarrow$}}
& \multirow{2}{*}{\makecell{Storage \\ (MB)$\downarrow$}}
& \multirow{2}{*}{\makecell{Train \\ (Sec)$\downarrow$}}
& \multirow{2}{*}{\makecell{Render \\ (FPS)$\uparrow$}} \\
& & \multicolumn{1}{c}{Coffee Martini} & \multicolumn{1}{c}{Cook Spinach} & \multicolumn{1}{c}{Cut Beef} & \multicolumn{1}{c}{Flame Salmon} & \multicolumn{1}{c}{Flame Steak} & \multicolumn{1}{c}{Sear Steak} & \multicolumn{1}{c|}{Average}
& & & & \\
\midrule
\multirow{9}{*}{\rotatebox{90}{Offline}}
& K-Planes$^{\dagger}$\cite{fridovich2023k} & 29.99 / 0.925 & 32.60 / 0.942 & 31.82 / 0.943 & 30.44 / 0.924 & 32.38 / 0.957 & 32.52 / 0.955 & 31.63 / 0.941 & - & 1.04 & 21.6 & 0.3 \\
& NeRFPlayer$^{\dagger}$\cite{song2023nerfplayer} & \textcolor{red}{\textbf{31.53}} / \textcolor{red}{\textbf{0.951}} & 30.56 / 0.929 & 29.35 / 0.908 & \textcolor{red}{\textbf{31.65}} / \textcolor{red}{\textbf{0.940}} & 31.93 / 0.950 & 29.13 / 0.908 & 30.69 / 0.931 & - & 17.1 & 72.0 & 0.05 \\
& 4DGaussians\cite{wu20244d}  & 30.10 / 0.935 & 31.15 / \textcolor{blue}{\underline{0.958}} & 32.60 / 0.951 & 30.21 / 0.934 & 33.49 / 0.961 & 32.64 / 0.961 & 31.70 / 0.950 & 192 & 0.29 & 7.92 & 23 \\
& Grid4D\cite{xu2024grid4d} & 28.69 / 0.919 & 32.90 / 0.957 & 33.61 / 0.958 & 29.86 / 0.930 & 32.98 / 0.963 & 33.49 / \textcolor{blue}{\underline{0.965}} & 31.92 / 0.949 & 202 & 0.16 & 16.4 & 127 \\
& STG\cite{li2024spacetime} & 28.61 / 0.916 & 33.18 / 0.952 & 33.52 / 0.954 & 29.48 / 0.918 & 33.64 / 0.960 & 33.89 / 0.961 & 32.05 / 0.944 & 434 & 0.67 & 15.1 & 149 \\
& 4DGS\cite{yang2023real} & 28.33 / 0.920 & 32.93 / 0.956 & 33.85 / \textcolor{blue}{\underline{0.959}} & 29.38 / 0.929 & 34.03 / 0.961 & 33.51 / 0.960 & 32.01 / 0.948 & 3125 & 20.9 & 66.9 & 119 \\
& E-D3DGS\cite{bae2024per} & 29.33 / 0.931 & 33.19 / 0.957 & 33.25 / 0.957 & 29.72 / 0.935 & 33.55 / 0.963 & 33.55 / 0.963 & 32.10 / 0.951 & 180 & 0.23 & 33.5 & 85 \\
& Ex4DGS\cite{lee2024fully} & 28.79 / 0.915 & 33.23 / 0.947 & 33.73 / 0.948 & 29.29 / 0.917 & 33.91 / 0.956 & 33.69 / 0.959 & 32.11 / 0.940 & 268 & 0.38 & 7.22 & 121 \\
\midrule
\multirow{9}{*}{\rotatebox{90}{Online}}
& StreamRF$^{\dagger}$\cite{li2022streaming} & 27.77 / 0.885 & 31.54 / 0.934 & 31.74 / 0.937 & 28.69 / 0.892 & 32.18 / 0.943 & 32.29 / 0.945 & 30.70 / 0.923 & - & 7.64 & 15.2 & 12 \\
& 3DGStream\cite{sun20243dgstream} & 27.75 / 0.917 & 32.22 / 0.957 & 33.67 / 0.957 & 28.61 / 0.924 & 33.47 / \textcolor{blue}{\underline{0.966}} & 33.39 / \textcolor{blue}{\underline{0.965}} & 31.69 / 0.948 & 326 & 8.14 & 8.11 & 245 \\
& 4DGC\cite{hu20254dgc} & 28.79 / 0.923 & 32.81 / 0.951 & 33.03 / 0.954 & 28.49 / 0.921 & 33.58 / 0.957 & 33.89 / \textcolor{blue}{\underline{0.965}} & 31.76 / 0.945 & 310 & 0.49 & 41.2 & 92\\
& HiCoM\cite{gao2024hicom} & 28.06 / 0.914 & 33.28 / 0.954 & 33.66 / 0.956 & 28.81 / 0.928 & 33.62 / 0.963 & 33.71 / 0.962 & 31.86 / 0.946 & 262 & 0.63 & 6.25 & 372 \\
& QUEEN-s\cite{girish2024queen} & 28.12 / 0.910 & 33.25 / 0.954 & 33.41 / 0.956 & 28.80 / 0.923 & 34.01 / 0.960 & 33.86 / 0.963 & 31.91 / 0.944 & 312 & 0.73 & 5.58 & 433\\
& QUEEN-l\cite{girish2024queen} & 28.27 / 0.913 & 33.49 / 0.955 & 33.87 / 0.957 & 29.35 / 0.926 & \textcolor{red}{\textbf{34.11}} / 0.961 & 33.97 / 0.962 & 32.18 / 0.946 & 315 & 0.79 & 7.43 & 417\\
& StreamSTGS\cite{ke2025streamstgs} & 28.83 / 0.913 & 33.35 / 0.953 & \textcolor{blue}{\underline{34.03}} / 0.956 & 29.68 / 0.924 & 33.76 / 0.957 & \textcolor{red}{\textbf{34.47}} / 0.959 & 32.35 / 0.944 & \textcolor{blue}{\underline{155}} & 0.17 & 25.5 & 199\\
& ReCon-GS$^{\dagger}$\cite{furecon} & \textcolor{blue}{\underline{30.14}} / \textcolor{blue}{\underline{0.938}} & \textcolor{blue}{\underline{33.54}} / \textcolor{red}{\textbf{0.961}} & 33.92 / \textcolor{red}{\textbf{0.966}} & \textcolor{blue}{\underline{30.43}} / 0.938 & 34.00 / \textcolor{red}{\textbf{0.969}} & 33.91 / \textcolor{red}{\textbf{0.966}} & \textcolor{blue}{\underline{32.66}} / \textcolor{red}{\textbf{0.956}} & 244 & 0.44 & 6.40 & 250 \\
& S²GS-fast (Ours) & 29.87 / 0.931 & 33.47 / 0.956 & 33.70 / 0.957 & 30.10 / 0.938 & 33.86 / 0.961 & 33.45 / 0.961 & 32.41 / 0.951 & \textcolor{red}{\textbf{101}} & \textcolor{blue}{\underline{0.12}} & \textcolor{red}{\textbf{2.26}} & \textcolor{red}{\textbf{484}} \\
& S²GS-full (Ours) & 30.07 / 0.933 & \textcolor{red}{\textbf{33.96}} / \textcolor{blue}{\underline{0.958}} & \textcolor{red}{\textbf{34.12}} / 0.958 & 30.26 / \textcolor{blue}{\underline{0.939}} & \textcolor{blue}{\underline{34.04}} / 0.962 & \textcolor{blue}{\underline{34.08}} / 0.962 & \textcolor{red}{\textbf{32.76}} / \textcolor{blue}{\underline{0.952}} & \textcolor{red}{\textbf{101}} & \textcolor{red}{\textbf{0.11}} & \textcolor{blue}{\underline{4.12}} & \textcolor{blue}{\underline{482}}\\
\bottomrule
\end{tabular}
\label{table_scene_comparison1}
\end{table*}

\begin{figure*}[ht]
  \centering
   \includegraphics[width=1.0\linewidth]{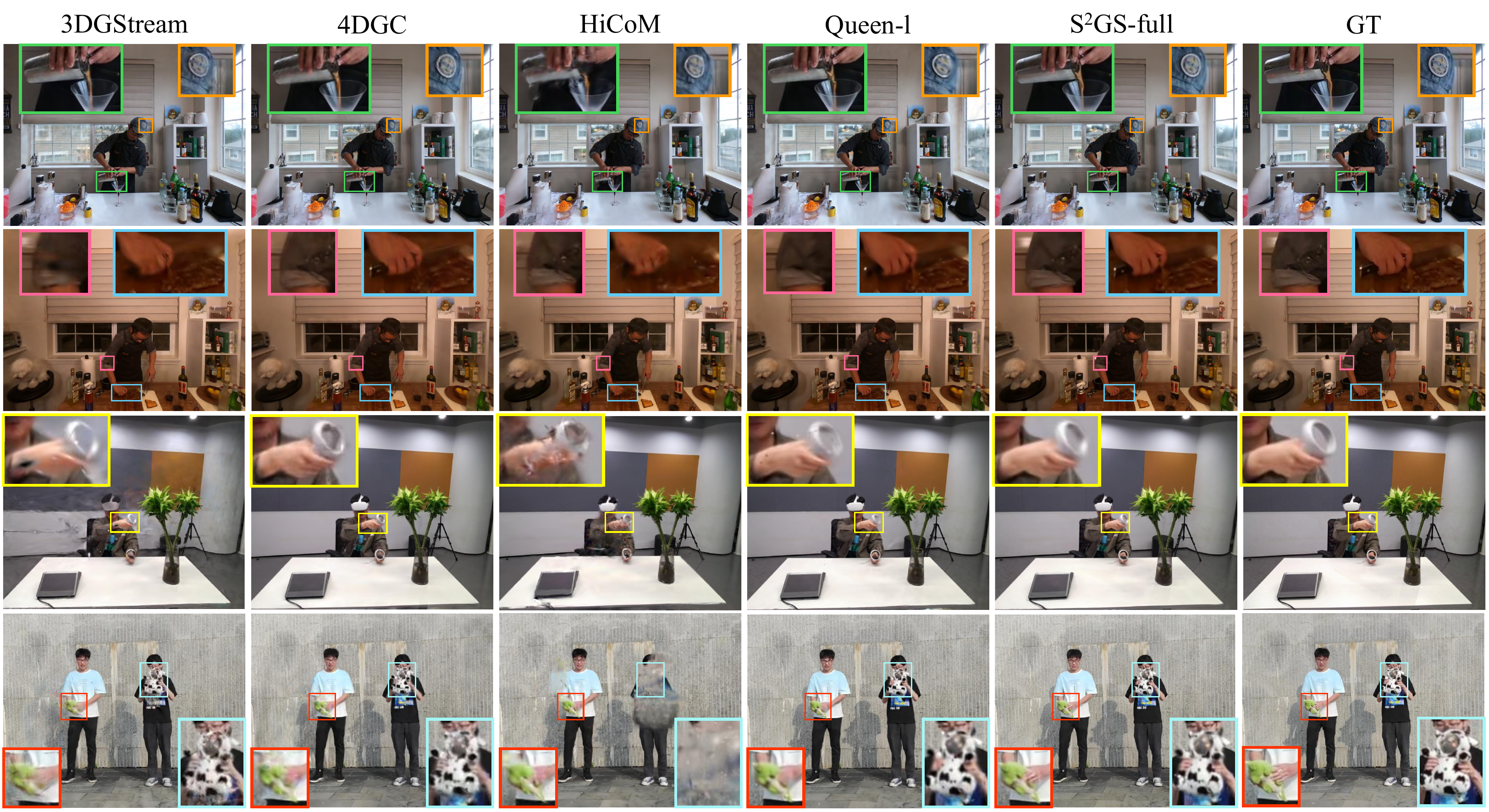}

   \caption{Qualitative comparison with SOTA methods \cite{sun20243dgstream, hu20254dgc, gao2024hicom, girish2024queen} on diverse datasets \cite{li2022neural, li2022streaming, lin2022efficient}. Our method captures fine-grained details presented in indoor and outdoor scenes, particularly for objects with intricate motions such as poured liquid, hands, knives, and stuffed toys.}
   \label{results1}
\end{figure*}

\subsection{Dataset}
We evaluate our method on three challenging FVV datasets:
\subsubsection{N3DV Dataset} N3DV \cite{li2022neural} includes six dynamic indoor scenes that exhibit various lighting conditions and intricate volumetric details. The videos were captured at a resolution of 2704 × 2028 and a frame rate of 30 FPS, with 18 to 21 camera views per scene. Following prior studies \cite{girish2024queen}, we downsampled the image by a factor of 2 for our experiments.
\subsubsection{Meet Room Dataset} Meet Room Dataset \cite{li2022streaming} was recorded using a multi-view system equipped with 13 synchronized cameras, covering three indoor scenes. The videos have a resolution of 1280 × 720 and a frame rate of 30 FPS.
\subsubsection{ENeRF Outdoor Dataset} ENeRF Outdoor Dataset \cite{lin2022efficient} contains long-sequence videos of complex human motions with deformable objects in an outdoor setting, captured by 18 cameras at a resolution of 1920 × 1080. We selected three 300-frame sequences with 6 different actors for experiments.

\subsection{Baseline, Metric, and Implementation}
\subsubsection{Baseline} We compare S²GS against several competitive online and offline FVV methods, including 4DGaussians\cite{wu20244d}, 4DGS\cite{yang2023real}, STG\cite{li2024spacetime}, Ex4DGS\cite{lee2024fully}, 3DGStream\cite{sun20243dgstream}, HiCoM\cite{gao2024hicom}, 4DGC\cite{hu20254dgc}, Queen\cite{girish2024queen}, and ReCon-GS\cite{furecon}. We evaluate three variants of S²GS—S²GS-full, S²GS-fast, and S²GS-edge—using 250 training epochs on the first frame and 20, 10, and 6 epochs, respectively, on each subsequent frame. In particular, S²GS-edge is tailored for deployment on resource-constrained edge-IoT devices.
\subsubsection{Metric} The quality of rendered images is assessed using PSNR, SSIM, and LPIPS with the VGG backbone. To evaluate streaming efficiency, we also report the average number of Gaussians, average storage size, average training time, and average rendering throughput per frame.
\subsubsection{Implementation Details} The main experiments for RQ-1, RQ2, and RQ-3 are conducted on an Ubuntu 22.04.5 LTS system equipped with an Intel Core i7-13700 CPU, an NVIDIA GeForce RTX 4090 GPU, and 64 GB of RAM. Deployment experiments are performed on the NVIDIA Jetson AGX Orin Developer Kit equipped with a 1.3 GHz NVIDIA Ampere architecture GPU, a 2.2 GHz ARM Cortex-A78AE processor, and 64GB of RAM. Additionally, in the implementation of S²GS, $\lambda_{1}$, $\lambda_{2}$, and $\eta$ in Eq.~\ref{eq_sparse_loss} are set to 0.01, 0.01, and 0.05, respectively. The $\operatorname{round}(\cdot)$ operation in Eq.~\ref{eq_ste} rounds soft gates to one decimal place to encourage sparse and discrete gates. All reported results are averaged over 3 independent runs to ensure statistical reliability.

\subsection{RQ-1 (Quality)}
\begin{figure*}[ht]
  \centering
   \includegraphics[width=1.0\linewidth]{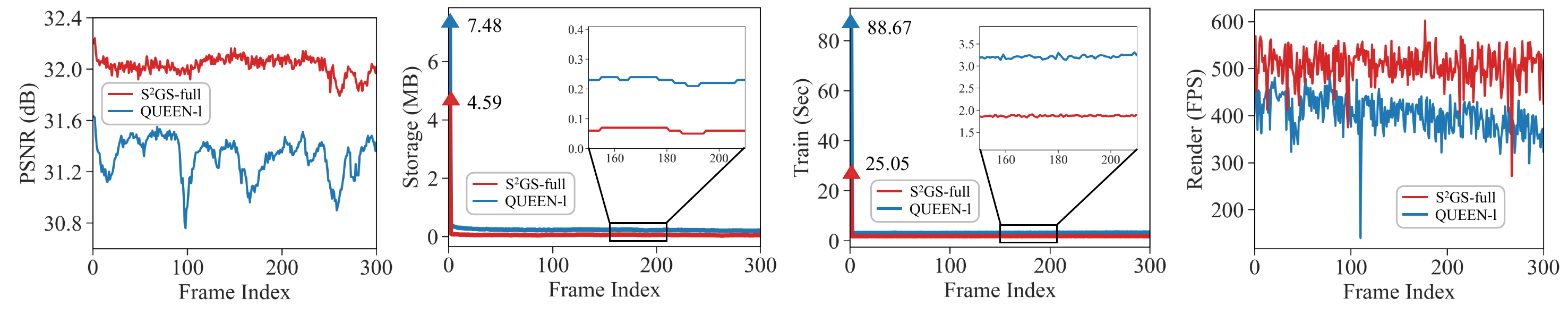}

   \caption{Trend comparison of PSNR, storage cost, training time, and rendering throughput for S²GS-full and QUEEN-l on the Vrheadset scene from the Meet Room dataset. The triangle ($\blacktriangle$) denotes the metric value at the first frame.}
   \label{results2}
\end{figure*}

Our method introduces a streaming octree structure that decouples multi-scale Gaussian residuals across different levels of dynamic variation. This design enables fine-scale Gaussians to capture dynamic details effectively, thereby improving overall rendering quality. Furthermore, the multi-level STE scheme mitigates gate collapse and produces fine-grained gate states, enabling the model to preserve sparse updates for intricate temporal dynamics. As shown in Figs.~\ref{results1} and ~\ref{results2}, Tabs.~\ref{table_scene_comparison1} and \ref{table_scene_comparison2}, we compare S²GS-full with state-of-the-art methods and demonstrate that our approach consistently achieves competitive or superior visual results on both indoor and outdoor scenes, particularly in regions exhibiting complex motions and intricate deformations. Notably, compared with QUEEN~\cite{girish2024queen} on ENeRF Outdoor, S$^2$GS-full achieves slightly lower PSNR (25.94 vs.~26.13~dB) but higher SSIM (0.863 vs.~0.843) and lower LPIPS (0.124 vs.~0.154), indicating a visual quality trade-off with slightly larger pixel-wise errors but better preservation of structural and perceptual details. More results can be found in the Supplementary Material.

\begin{table*}[ht]
\centering
\caption{Quantitative comparison on Meet Room and ENeRF Outdoor datasets. For all methods, metrics are averaged across all frames. Results for methods marked with $^{\dagger}$ are taken from the original papers. All experiments are conducted on an NVIDIA RTX 4090 GPU. \textcolor{red}{\textbf{Red}} and \textcolor{blue}{\underline{Blue}} indicate the best and second best, respectively, within each category.}
\footnotesize
\setlength{\tabcolsep}{4.0pt}
\renewcommand{\arraystretch}{1.05}
\begin{tabular}{
l|  
l|  
ccccccc|  
ccccccc  
}
\toprule
& \multirow{3}{*}{Methods}
& \multicolumn{7}{c|}{Meet Room Dataset} 
& \multicolumn{7}{c}{ENeRF Outdoor Dataset} \\

& & \multicolumn{1}{c}{\makecell{PSNR \\ (dB)$\uparrow$}} & \multicolumn{1}{c}{\makecell{SSIM \\ $\uparrow$}} & \multicolumn{1}{c}{\makecell{LPIPS \\ $\downarrow$}} & \multicolumn{1}{c}{\makecell{\#Gaussians\\ (k)$\downarrow$}} & \multicolumn{1}{c}{\makecell{Storage \\ (MB)$\downarrow$}} & \multicolumn{1}{c}{\makecell{Train \\ (Sec)$\downarrow$}} & \multicolumn{1}{c|}{\makecell{Render \\ (FPS)$\uparrow$}} & \multicolumn{1}{c}{\makecell{PSNR \\ (dB)$\uparrow$}} & \multicolumn{1}{c}{\makecell{SSIM \\ $\uparrow$}} & \multicolumn{1}{c}{\makecell{LPIPS \\ $\downarrow$}} & \multicolumn{1}{c}{\makecell{\#Gaussians\\ (k)$\downarrow$}} & \multicolumn{1}{c}{\makecell{Storage \\ (MB)$\downarrow$}} & \multicolumn{1}{c}{\makecell{Train \\ (Sec)$\downarrow$}} & \multicolumn{1}{c}{\makecell{Render \\ (FPS)$\uparrow$}}\\
\midrule
\multirow{10}{*}{\rotatebox{90}{Online}}
& 3DGStream\cite{sun20243dgstream} & 29.30 & 0.942 & 0.188  & 185 & 4.10 & 4.77 & 260 & 25.91 & 0.856 & 0.129 & 2182 & 8.77 & 12.96 & 117 \\
& 4DGC\cite{hu20254dgc} & 30.41 & 0.947 & 0.180 & 208 & 0.77 & 24.52 & 155 & 24.52 & 0.845 & 0.135 & 1275 & 0.93 & 37.71 & 102 \\
& HiCoM\cite{gao2024hicom} & 29.57 & 0.944 & 0.182 & 152 & 0.39 & 3.91 & 377 & 22.12 & 0.749 & 0.280 & 761 & 1.74 & 4.43 & 385\\
& QUEEN-s\cite{girish2024queen} & 31.95 & 0.960 & 0.159 & 133 & 0.23 & 2.92 & 414 & \textcolor{red}{\textbf{26.21}} & 0.839 & 0.177 & 282 & 0.95 & 2.94 & 411\\
& QUEEN-l\cite{girish2024queen} & \textcolor{blue}{\underline{32.21}} & 0.962 & 0.152 & 135 & 0.29 & 3.51 & 397 & \textcolor{blue}{\underline{26.13}} & 0.843 & 0.154 & 354 & 1.32 & 3.82 & 358\\
& StreamSTGS\cite{ke2025streamstgs} & 27.15 & 0.914 & 0.231 & \textcolor{blue}{\underline{84}} & 0.13 & 9.19 & 220 & 25.83 & 0.759 & 0.193 & \textcolor{blue}{\underline{164}} & \textcolor{blue}{\underline{0.22}} & 13.67 & 191 \\
& ReCon-GS$^{\dagger}$\cite{furecon} & 30.84 & 0.955 & 0.163 & - & 0.30 & 3.86 & 256 & - & - & - & - & - & - & -\\
& S²GS-fast (Ours) & 32.20 & \textcolor{blue}{\underline{0.963}} & \textcolor{blue}{\underline{0.149}} & \textcolor{red}{\textbf{64}} & \textcolor{blue}{\underline{0.10}} & \textcolor{red}{\textbf{1.13}} & \textcolor{blue}{\underline{498}} & 26.07 & \textcolor{blue}{\underline{0.861}} & \textcolor{blue}{\underline{0.125}} & \textcolor{red}{\textbf{138}} & 0.25 & \textcolor{red}{\textbf{1.22}} & \textcolor{blue}{\underline{539}}\\
& S²GS-full (Ours) & \textcolor{red}{\textbf{32.46}} & \textcolor{red}{\textbf{0.964}} & \textcolor{red}{\textbf{0.147}} & \textcolor{red}{\textbf{64}} & \textcolor{red}{\textbf{0.08}} & \textcolor{blue}{\underline{1.96}} & \textcolor{red}{\textbf{513}} & 25.94 & \textcolor{red}{\textbf{0.863}} & \textcolor{red}{\textbf{0.124}} & \textcolor{red}{\textbf{138}} & \textcolor{red}{\textbf{0.20}} & \textcolor{blue}{\underline{2.28}} & \textcolor{red}{\textbf{541}}\\
\bottomrule
\end{tabular}
\label{table_scene_comparison2}
\end{table*}

\subsection{RQ-2 (Efficiency)}

\subsubsection{Model Complexity}
As shown in Tabs.~\ref{table_scene_comparison1} and \ref{table_scene_comparison2}, S²GS significantly reduces the number of Gaussian primitives required for FVV streaming, thereby lowering model complexity during both training and rendering. These results highlight the effectiveness of the proposed streaming octree representation, which reorganizes dynamic scenes using a hierarchical and compact Gaussian distribution.

\subsubsection{Storage}
By reducing model complexity, our method substantially lowers the storage cost required for streaming FVV transmission over IoT infrastructure. Moreover, the structured gating mechanism further compresses storage usage by selectively learning and storing sparse residual updates and discrete gates. As shown in Tabs.~\ref{table_scene_comparison1} and \ref{table_scene_comparison2}, S²GS consistently outperforms existing quantization-based approaches \cite{girish2024queen, hu20254dgc} across all evaluated datasets, highlighting its superior storage efficiency. The storage cost trend in Fig.~\ref{results2} further demonstrates that, compared with QUEEN \cite{girish2024queen}, our method achieves lower storage consumption for both the initial key frame and subsequent online streaming frames.

\subsubsection{Online Training and Rendering}
Another key contribution of S²GS lies in accelerating online training and rendering for FVV streaming under resource constraints. As reported in Tabs.~\ref{table_scene_comparison1} and \ref{table_scene_comparison2}, our method captures competitive dynamic details while consistently achieving the shortest per-frame training time and the highest rendering throughput. Specifically, on the N3DV dataset, S²GS-full reduces the per-frame training time to under 5 seconds, while S²GS-fast further lowers it to 2.26 seconds, all while maintaining a rendering throughput exceeding 480 FPS. The performance trajectories in Fig.~\ref{results2} further show that, compared with QUEEN~\cite{girish2024queen}, our method consistently reduces per-frame optimization time and improves rendering throughput, supporting its practicality for efficient FVV streaming in edge-IoT scenarios.

\begin{table}[t]
\centering
\caption{Ablation study on the N3DV dataset. \textcolor{red}{\textbf{Red}} and \textcolor{blue}{\underline{Blue}} indicate the best and second best, respectively, within each category.}
\footnotesize
\setlength{\tabcolsep}{3.0pt}
\renewcommand{\arraystretch}{1.1}
\begin{tabular}{
l|  
cccccc
}
\toprule
Incremental Design & \multicolumn{1}{c}{\makecell{PSNR \\ (dB)$\uparrow$}} & \multicolumn{1}{c}{\makecell{\#Gaussians \\ (k)$\downarrow$}} & \multicolumn{1}{c}{\makecell{\#Active \\Gates $\downarrow$}} & \multicolumn{1}{c}{\makecell{Storage \\ (MB)$\downarrow$}} & \multicolumn{1}{c}{\makecell{Train \\ (Sec)$\downarrow$}} & \multicolumn{1}{c}{\makecell{Render \\ (FPS)$\uparrow$}}\\
\midrule

Baseline & 29.61 & 412 & 410k & 3.09 & 13.7 & 199 \\

\hspace{0.6em}+ STE  & 31.89 & 352 & 107k & 2.37 & 12.9 & 221 \\

\hspace{0.6em}+ Sparse Reg. & 32.25 & \textcolor{blue}{\underline{331}} & 16k & 1.10 & 12.7 & 234 \\

\hspace{0.6em}+ Stream. Octree & \textcolor{blue}{\underline{32.71}} & \textcolor{red}{\textbf{101}} & \textcolor{blue}{\underline{2.4k}} & \textcolor{blue}{\underline{0.29}} & \textcolor{blue}{\underline{4.24}} & \textcolor{red}{\textbf{485}}\\

\hspace{0.6em}+ HFP (S²GS-full) & \textcolor{red}{\textbf{32.76}} & \textcolor{red}{\textbf{101}} & \textcolor{red}{\textbf{679}} & \textcolor{red}{\textbf{0.11}} & \textcolor{red}{\textbf{4.12}} & \textcolor{blue}{\underline{482}}\\

\bottomrule

\end{tabular}
\label{table_ablation1}
\end{table}

\begin{table*}[t]
\centering
\caption{2 $\times$ 2 factorial ablation study of structure and sparsity. Structure denotes the proposed streaming octree with HFP, and sparsity denotes STE with sparse regularization. \textcolor{red}{\textbf{Red}} and \textcolor{blue}{\underline{Blue}} indicate the best and second-best results, respectively.}
\footnotesize
\setlength{\tabcolsep}{4.0pt}
\renewcommand{\arraystretch}{1.1}
\begin{tabular}{cc|cc|cccccc|cccccc}
\toprule
\multicolumn{2}{c|}{Structure} & \multicolumn{2}{c|}{Sparsity} & \multicolumn{6}{c|}{MeetRoom Dataset} & \multicolumn{6}{c}{ENeRF Outdoor Dataset} \\
\cmidrule(r){1-2}\cmidrule(r){3-4}\cmidrule(r){5-10}\cmidrule(r){11-16}
\makecell{Stream. \\ Octree} 
& \makecell{HFP}
& \makecell{STE} 
& \makecell{Sparse\\ Reg.} 
& \makecell{PSNR \\ (dB)$\uparrow$}
& \makecell{\#Gaussians \\ (k)$\downarrow$}
& \makecell{\#Active \\ Gates$\downarrow$}
& \makecell{Storage \\ (MB)$\downarrow$}
& \makecell{Train \\ (Sec)$\downarrow$}
& \makecell{Render \\ (FPS)$\uparrow$}
& \makecell{PSNR \\ (dB)$\uparrow$}
& \makecell{\#Gaussians \\ (k)$\downarrow$}
& \makecell{\#Active \\ Gates$\downarrow$}
& \makecell{Storage \\ (MB)$\downarrow$}
& \makecell{Train \\ (Sec)$\downarrow$}
& \makecell{Render \\ (FPS)$\uparrow$} \\
\midrule
- & - & - & - & 31.69 & 129 & 129k & 0.67 & 2.86 & 383 & 25.12 & 439 & 398k & 4.17 & 4.03 & 285  \\

 - & - & $\checkmark$ & $\checkmark$ & 31.86 & \textcolor{blue}{\underline{127}} & \textcolor{blue}{\underline{13k}} & \textcolor{blue}{\underline{0.22}} & 2.73 & 387 & 25.69 & \textcolor{blue}{\underline{382}} & 51k & \textcolor{blue}{\underline{1.41}} & 3.66 & 299 \\

$\checkmark$ & $\checkmark$ & - & - & \textcolor{blue}{\underline{32.33}} & \textcolor{red}{\textbf{64}} & 17k & 0.38 & \textcolor{blue}{\underline{2.18}} & \textcolor{blue}{\underline{496}} & \textcolor{blue}{\underline{25.80}} & \textcolor{red}{\textbf{138}} & \textcolor{blue}{\underline{39k}} & 1.77 & \textcolor{blue}{\underline{2.45}} & \textcolor{blue}{\underline{532}}\\
$\checkmark$ & $\checkmark$ & $\checkmark$ & $\checkmark$  & \textcolor{red}{\textbf{32.46}} & \textcolor{red}{\textbf{64}} & \textcolor{red}{\textbf{401}} & \textcolor{red}{\textbf{0.08}} & \textcolor{red}{\textbf{1.96}} & \textcolor{red}{\textbf{513}} & \textcolor{red}{\textbf{25.94}} & \textcolor{red}{\textbf{138}} & \textcolor{red}{\textbf{1.9k}} & \textcolor{red}{\textbf{0.20}} & \textcolor{red}{\textbf{2.28}} & \textcolor{red}{\textbf{541}} \\

\bottomrule
\end{tabular}
\label{table_ablation2}
\end{table*}

\begin{table}[t]
\centering
\caption{Propagation Strategy Analysis on the ENeRF Outdoor Dataset. \textcolor{red}{\textbf{Red}} and \textcolor{blue}{\underline{Blue}} indicate the best and second best, respectively. FT denotes fixed-threshold pruning.}
\footnotesize
\setlength{\tabcolsep}{3.0pt}
\renewcommand{\arraystretch}{1.1}
\begin{tabular}{
l|  
cccccc
}
\toprule
Configuration & \multicolumn{1}{c}{\makecell{PSNR \\ (dB)$\uparrow$}} & \multicolumn{1}{c}{\makecell{LPIPS \\ $\downarrow$}} & \multicolumn{1}{c}{\makecell{\#Active \\Gates $\downarrow$}} & \multicolumn{1}{c}{\makecell{Storage \\ (MB)$\downarrow$}} & \multicolumn{1}{c}{\makecell{Train \\ (Sec)$\downarrow$}} & \multicolumn{1}{c}{\makecell{Render \\ (FPS)$\uparrow$}}\\
\midrule
HFP (S²GS-full) & \textcolor{red}{\textbf{25.94}} & \textcolor{red}{\textbf{0.124}} & 1.9k & 0.20 & \textcolor{red}{\textbf{2.28}} & \textcolor{blue}{\underline{541}} \\
FT, $\tau=0.50$ & \textcolor{blue}{\underline{25.82}} & \textcolor{blue}{\underline{0.128}} & 1.8k & 0.19 & 2.35 & 537 \\
FT, $\tau=0.80$ & 25.72 & 0.133 & 1.6k & 0.18 & 2.39 & 526 \\
FT, $\tau=0.90$ & 25.33 & 0.141 & \textcolor{blue}{\underline{1.2k}} & \textcolor{blue}{\underline{0.15}} & 2.37 & 539 \\
FT, $\tau=0.95$ & 25.12 & 0.150 & \textcolor{red}{\textbf{707}} & \textcolor{red}{\textbf{0.13}} & \textcolor{blue}{\underline{2.32}} & \textcolor{red}{\textbf{544}} \\
\bottomrule

\end{tabular}
\label{table_ablation3}
\end{table}

\begin{table}[t]
\centering
\caption{Comparison of discrete gating strategies on the N3DV dataset. \textcolor{red}{\textbf{Red}} indicates the best result within each category. ML denotes the multi-level scheme.}
\footnotesize
\setlength{\tabcolsep}{3.0pt}
\renewcommand{\arraystretch}{1.1}
\begin{tabular}{
l|  
cccccc
}
\toprule
Configuration & \multicolumn{1}{c}{\makecell{PSNR \\ (dB)$\uparrow$}} & \multicolumn{1}{c}{\makecell{SSIM \\ $\uparrow$}} & \multicolumn{1}{c}{\makecell{\#Active \\Gates $\downarrow$}} & \multicolumn{1}{c}{\makecell{Storage \\ (MB)$\downarrow$}} & \multicolumn{1}{c}{\makecell{Train \\ (Sec)$\downarrow$}} & \multicolumn{1}{c}{\makecell{Render \\ (FPS)$\uparrow$}}\\
\midrule

STE: Binary & 29.85 & 0.934 & \textcolor{red}{\textbf{2}} & \textcolor{red}{\textbf{0.06}} & \textcolor{red}{\textbf{4.05}} & 478 \\

STE: ML (S²GS-full)  & \textcolor{red}{\textbf{32.76}} & \textcolor{red}{\textbf{0.952}} & 679 & 0.11 & 4.12 & \textcolor{red}{\textbf{482}} \\
\bottomrule

\end{tabular}
\label{table_ablation4}
\end{table}

\begin{figure}[t]
  \centering
   \includegraphics[width=1.0\linewidth]{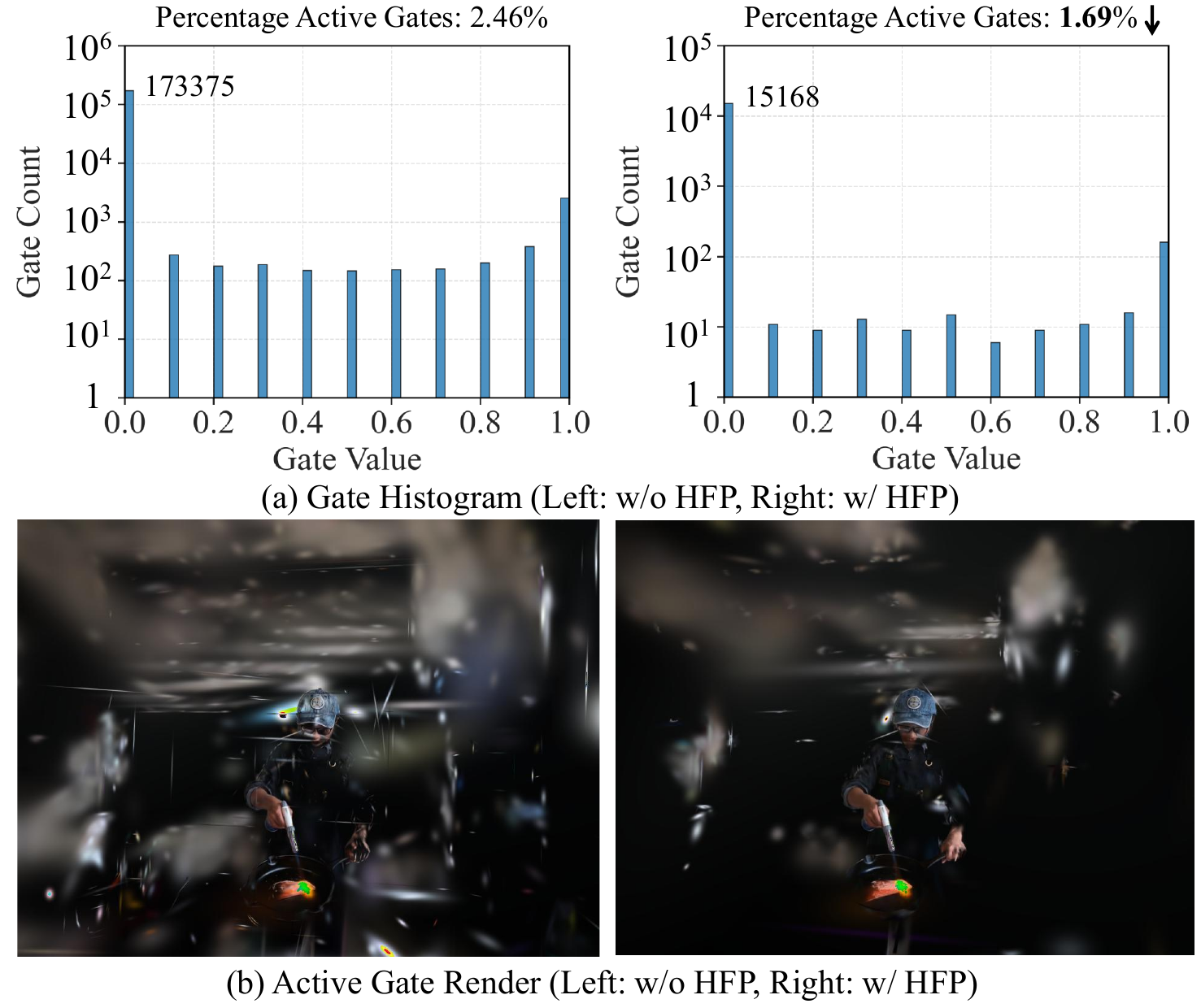}

   \caption{Effect of hierarchical feature propagation (HFP) on the Flame Salmon scene. (a) With HFP, the number of sparse active gates is reduced by nearly one order of magnitude at a lower ratio. (b) HFP effectively identifies dynamic regions while suppressing noise.}
   \label{results3}
\end{figure}

\subsection{RQ-3 (Ablation Study)}
We conduct ablation studies from three perspectives. First, we progressively build the full S²GS framework to evaluate the contribution of each component. Second, we disentangle the effects of the structure and sparsity-related modules to verify whether they are complementary. Third, we validate two key design choices, namely the multi-level STE and the proposed HFP strategy. We use S$^2$GS-full as the complete model and a vanilla GS-based FVV model without residual sparsification or the octree structure as the baseline. To ensure controlled comparisons, all ablation variants are optimized using the same fixed budget of 20 epochs per subsequent frame.

\subsubsection{Component Analysis} As shown in Tab. \ref{table_ablation1}, each component contributes positively to the final S²GS framework.

\textit{Straight-Through Estimator (STE):} Starting from the baseline, introducing STE reduces the number of active gates to approximately 1/4 of the original count, thereby lowering storage requirements and per-frame optimization time. This indicates that STE effectively sparsifies residual updates.

\textit{Sparse Regularization:} As reported in Tab. \ref{table_ablation1}, introducing the regularization term substantially reduces the number of active gates, thereby reducing storage overhead. Moreover, as with STE, rendering quality consistently improves because the regularization helps stabilize residual optimization and suppress excessively dynamic or noisy updates in FVV streaming.

\textit{Streaming Octree:} The proposed streaming octree substantially reduces model complexity by globally reorganizing Gaussian residuals into a compact hierarchy. This design reduces the number of optimized Gaussians and active gates, thereby reducing storage footprints and per-frame optimization cost. Meanwhile, it improves visual quality by capturing multi-scale and fine-grained dynamic details. For deeper analysis of the octree structure, please check Appendix \ref{Streaming Octree Representation} for details.

\textit{Hierarchical Feature Propagation (HFP):} To assess its impact on gate learning, we ablate the HFP module. As reported in Tab. \ref{table_ablation1}, HFP reduces both storage cost and training time. By propagating features hierarchically, gate learning is performed on a smaller and more stable set of root-level regions rather than leaf-level Gaussians. Consequently, as shown in Fig.~\ref{results3}, this design yields nearly an order-of-magnitude fewer active gates, sparser residual updates (1.69\% vs.\ 2.46\%), and more robust dynamic region identification with reduced noise.

\subsubsection{Complementary Effects of Structure and Sparsity}
We conduct a $2 \times 2$ factorial ablation study to evaluate the complementary effects of structural modeling (streaming octree with HFP) and sparsity modeling (STE with sparse regularization). As shown in Table~\ref{table_ablation2}, sparsity alone substantially reduces active gates and storage while slightly improving PSNR, whereas structure alone reduces model complexity and substantially improves PSNR and rendering throughput. When combined, the two factors produce synergistic gains: extremely sparse active gates, minimal storage and training time, and maintained or modestly improved PSNR. These results confirm that structured representation and sparse residual updates are complementary rather than redundant: the former enhances hierarchical propagation and representation efficiency, while the latter suppresses unnecessary updates and enforces sparsity in temporal dynamics.

\subsubsection{Additional Design Analysis} We conduct additional experiments to validate the design choices of HFP and STE.

\begin{figure}[t]
  \centering
   \includegraphics[width=1.0\linewidth]{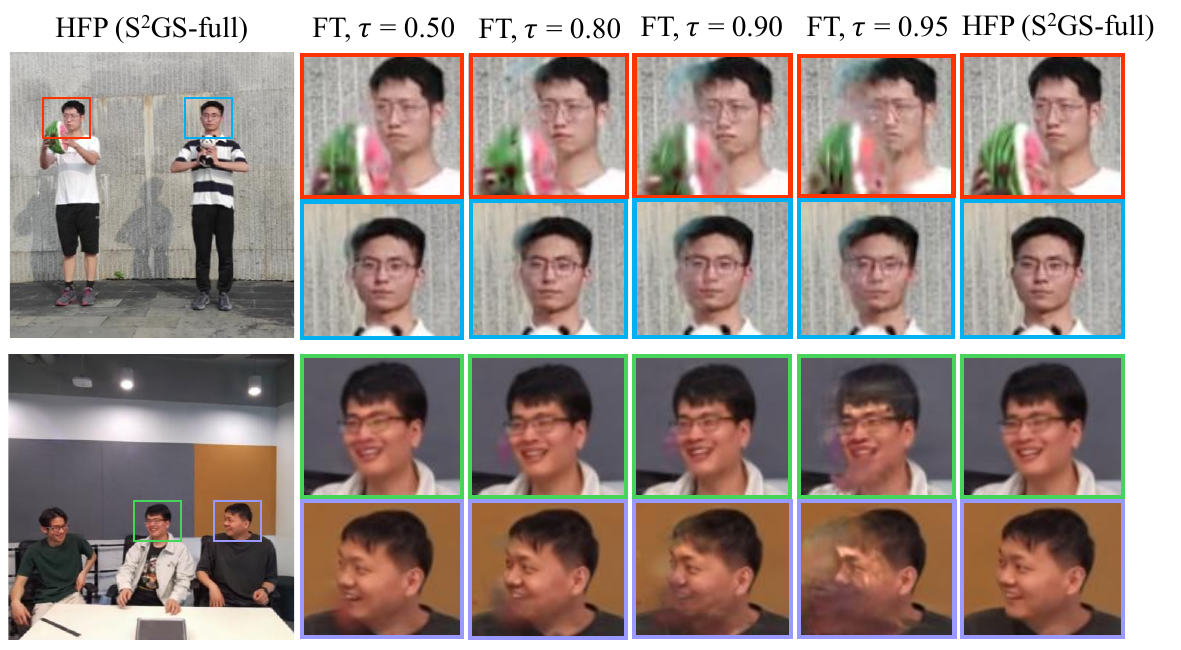}

   \caption{Qualitative comparison of different propagation strategies. FT-based variants progressively blur high-motion areas as $\tau$ increases, whereas HFP preserves fine details and sharpness. Results indicate that HFP effectively retains informative dynamic responses, particularly in regions with subtle but meaningful local motion.}
   \label{FT}
\end{figure}

\textit{Propagation Strategy Analysis:} Tab.~\ref{table_ablation3} compares the proposed HFP with fixed-threshold (FT) pruning under different threshold values $\tau$. HFP achieves the best overall reconstruction quality. In contrast, FT-based variants exhibit a clear degradation in reconstruction quality as $\tau$ increases, especially in highly dynamic regions (see Fig.~\ref{FT} for qualitative results). These observations suggest that FT-based propagation tends to remove informative dynamic responses, especially those associated with weak but meaningful local motion. In contrast, HFP preserves effective activations (1.9k active gates) and consistently achieves better fidelity than FT settings, indicating that the HFP-based S²GS framework remains sensitive to complex and salient local dynamics. The implementation details of the FT-based method can be found in Appendix \ref{Propagation Strategy Analysis}. 

\textit{Binary STE vs. Multi-Level (ML) STE:} Tab. \ref{table_ablation4} compares two discrete gating strategies, namely binary STE and the proposed multi-level STE, within the same S²GS framework. Binary STE significantly degrades reconstruction quality, reducing PSNR/SSIM from 32.76/0.952 to 29.85/0.934. The average number of active gates per frame is 2, indicating a severe collapse to an excessively sparse solution that suppresses necessary residual updates, as illustrated in Fig. \ref{ste}. In contrast, multi-level STE provides finer discrete granularity, which better balances sparsity and reconstruction fidelity. As a result, it preserves a richer set of effective updates (679 active gates) and achieves markedly better visual quality with nearly unchanged training and rendering cost.

\begin{table*}[t]
\centering
\caption{Quantitative comparison on the Jetson AGX Orin with the multi-resolution N3DV dataset. Energy per frame is computed as average power multiplied by per-frame training time. \textcolor{red}{\textbf{Red}} and \textcolor{blue}{\underline{Blue}} indicate the best and second best, respectively.}
\footnotesize
\setlength{\tabcolsep}{4.5pt}
\renewcommand{\arraystretch}{1.0}
\begin{tabular}{
l|  
l|  
ccccccc|
ccccc  
}
\toprule
\multirow{3}{*}{Resolution}
& \multirow{3}{*}{Methods}
& \multicolumn{7}{c|}{Quality \& Efficiency} 
& \multicolumn{5}{c}{Resource Consumption}\\

& & \multicolumn{1}{c}{\makecell{PSNR \\ (dB)$\uparrow$}} & \multicolumn{1}{c}{\makecell{SSIM \\ $\uparrow$}} & \multicolumn{1}{c}{\makecell{LPIPS \\ $\downarrow$}} & \multicolumn{1}{c}{\makecell{\#Gaussians\\ (k)$\downarrow$}} & \multicolumn{1}{c}{\makecell{Storage \\ (MB)$\downarrow$}} & \multicolumn{1}{c}{\makecell{Train \\ (Sec)$\downarrow$}} & \multicolumn{1}{c|}{\makecell{Render \\ (FPS)$\uparrow$}} & \multicolumn{1}{c}{\makecell{CPU\\ (\%)$\downarrow$}} & \multicolumn{1}{c}{\makecell{GPU MEM \\ (GB) $\downarrow$}} & \multicolumn{1}{c}{\makecell{RAM \\ (GB)$\downarrow$}} & \multicolumn{1}{c}{\makecell{Power \\ (W)$\downarrow$}} & \multicolumn{1}{c}{\makecell{Energy \\ (J)$\downarrow$}} \\
\midrule
\multirow{4}{*}{1/4}
& 3DGStream\cite{sun20243dgstream} & \textcolor{red}{\textbf{33.21}} & 0.953 & 0.095 & 386 & 8.21 & 63.28 & 21.22 & \textcolor{blue}{\underline{38.23}} & 1.82 & \textcolor{red}{\textbf{0.78}} & 14.56 & 921.36 \\
& HiCoM\cite{gao2024hicom} & 31.25 & 0.950 & \textcolor{blue}{\underline{0.085}} & \textcolor{blue}{\underline{194}} & \textcolor{blue}{\underline{0.49}} & 45.61 & 24.81 & 208.0 & \textcolor{blue}{\underline{1.53}} & 1.34 & \textcolor{blue}{\underline{13.53}} & 617.10 \\
& QUEEN-s\cite{girish2024queen} & 32.40 & \textcolor{blue}{\underline{0.955}} & \textcolor{blue}{\underline{0.085}} & 260 & 0.60 & \textcolor{blue}{\underline{32.27}} & \textcolor{blue}{\underline{32.35}} & \textcolor{red}{\textbf{25.86}} & 1.91 & \textcolor{blue}{\underline{1.29}} & 14.29 & \textcolor{blue}{\underline{461.14}} \\
& S²GS-edge (Ours) & \textcolor{blue}{\underline{33.07}} & \textcolor{red}{\textbf{0.958}} & \textcolor{red}{\textbf{0.083}} & \textcolor{red}{\textbf{82}} & \textcolor{red}{\textbf{0.14}} & \textcolor{red}{\textbf{18.61}} & \textcolor{red}{\textbf{49.73}} & 41.98 & \textcolor{red}{\textbf{1.42}} & 1.32 & \textcolor{red}{\textbf{11.21}} & \textcolor{red}{\textbf{208.62}}\\
\midrule
\multirow{4}{*}{1/8}
& 3DGStream\cite{sun20243dgstream} & 32.48 & 0.952 & 0.063 & 397 & 8.17 & 38.90 & 28.77 & \textcolor{red}{\textbf{40.64}} & 1.71 & \textcolor{red}{\textbf{0.74}} & 15.55 & 604.90 \\
& HiCoM\cite{gao2024hicom} & 31.24 & 0.956 & 0.051 & \textcolor{blue}{\underline{136}} & \textcolor{blue}{\underline{0.34}} & 21.70 & 46.53 & 227.7 & \textcolor{blue}{\underline{1.13}} & \textcolor{blue}{\underline{1.10}} & \textcolor{blue}{\underline{13.31}} & 288.83\\
& QUEEN-s\cite{girish2024queen} & \textcolor{blue}{\underline{32.90}} & \textcolor{blue}{\underline{0.967}} & \textcolor{blue}{\underline{0.038}} & 202 & 0.48 & \textcolor{blue}{\underline{17.04}} & \textcolor{blue}{\underline{52.06}} & \textcolor{blue}{\underline{44.70}} & 2.17 & 1.26 & 14.08 & \textcolor{blue}{\underline{239.92}}\\
& S²GS-edge (Ours) & \textcolor{red}{\textbf{33.26}} & \textcolor{red}{\textbf{0.968}} & \textcolor{red}{\textbf{0.036}} & \textcolor{red}{\textbf{50}} & \textcolor{red}{\textbf{0.10}} & \textcolor{red}{\textbf{8.622}} & \textcolor{red}{\textbf{67.94}} & 66.74 & \textcolor{red}{\textbf{1.08}} & 1.24 & \textcolor{red}{\textbf{11.18}} & \textcolor{red}{\textbf{96.39}}\\
\midrule
\multirow{4}{*}{{1/16}}
& 3DGStream\cite{sun20243dgstream} & 31.17 & 0.944 & 0.065 & 407 & 8.16 & 35.29 & 30.56 & \textcolor{red}{\textbf{42.13}} & 1.72 & \textcolor{red}{\textbf{0.73}} & 15.41 & 543.82 \\
& HiCoM\cite{gao2024hicom} & 31.65 & 0.958 & 0.047 & \textcolor{blue}{\underline{103}} & \textcolor{blue}{\underline{0.26}} & 18.72 & 49.11 & 184.5 & \textcolor{red}{\textbf{0.89}} & 1.25 & \textcolor{blue}{\underline{12.53}} & 234.56\\
& QUEEN-s\cite{girish2024queen} & \textcolor{blue}{\underline{33.44}} & \textcolor{blue}{\underline{0.977}} & \textcolor{blue}{\underline{0.020}} & 176 & 0.43 & \textcolor{blue}{\underline{10.23}} & \textcolor{blue}{\underline{61.06}} & \textcolor{blue}{\underline{53.19}} & 1.22 & \textcolor{blue}{\underline{1.15}} & 14.48 & \textcolor{blue}{\underline{148.13}} \\
& S²GS-edge (Ours) & \textcolor{red}{\textbf{33.57}} & \textcolor{red}{\textbf{0.978}} & \textcolor{red}{\textbf{0.018}} & \textcolor{red}{\textbf{24}} & \textcolor{red}{\textbf{0.06}} & \textcolor{red}{\textbf{5.612}} & \textcolor{red}{\textbf{86.66}} & 87.43 & \textcolor{blue}{\underline{0.92}} & 1.17 & \textcolor{red}{\textbf{11.46}} & \textcolor{red}{\textbf{64.31}}\\
\bottomrule
\end{tabular}
\label{table_edge}
\end{table*}

\subsection{RQ-4 (Deployability)}

We conduct deployment experiments on the NVIDIA Jetson AGX Orin Developer Kit, a leading industrial edge-IoT platform. We assess visual quality, efficiency, and resource consumption of four baseline models on the multi-resolution N3DV dataset, as summarized in Tab. \ref{table_edge}. Resource usage metrics are collected using the $\operatorname{jtop}$ monitoring tool on Jetson \cite{jetson_stats}, including averaged CPU utilization, GPU memory usage, RAM usage, and power consumption. We further compute the reconstruction energy per frame as the average power consumption multiplied by the per-frame training time. This metric provides a practical estimate of the dominant energy cost of streaming reconstruction, as the measured processing time is dominated by per-frame optimization. 

Several key observations can be drawn. \textit{Quality:} S²GS-edge consistently achieves higher visual quality by adaptively adjusting the octree resolution, demonstrating robustness in FVV streaming with multi-resolution inputs. \textit{Timeliness:} S²GS-edge significantly outperforms other efficient baselines in both per-frame training time and rendering throughput. Specifically, it achieves the per-frame training time below 6 seconds and rendering throughput above 85 FPS on the N3DV dataset at 1/16 resolution, satisfying fast training and real-time rendering requirements. \textit{Resource Consumption:} Thanks to its reduced model complexity and efficient per-frame training, S²GS-edge exhibits the lowest power consumption, per-frame energy cost, and GPU memory footprint, while maintaining competitive RAM usage compared to other baselines. In summary, S²GS-edge demonstrates strong visual quality, high timeliness, and low resource consumption, making it well-suited for resource-constrained edge-IoT environments.

\subsection{RQ-5 (Case Study)}
We develop an FVV streaming system based on a telerobotic platform, as illustrated in Fig. \ref{testbed}. Immersive digital twins of human operators are telepresented online to record and demonstrate the operation procedures of robotic arms.
\subsubsection{Physical Testbed} 
The physical testbed is designed to assess whether S$^2$GS can support FVV streaming under real-world acquisition and resource constraints. The FVV capture setup consists of 9 industrial cameras connected via 3 USB 3.0 hubs. Multi-view videos are captured at a resolution of 320 × 240 and synchronized using ROS 2 on a cloud server. The center camera is selected as the test view, and the first frame is processed using COLMAP \cite{schoenberger2016sfm, schoenberger2016mvs} to estimate the initial point clouds and camera poses. Based on these initializations, the FVV streaming stage, including per-frame optimization and rendering, is performed on the NVIDIA Jetson AGX Orin.
\subsubsection{Experimental Results} 
\begin{figure}[t]
  \centering
   \includegraphics[width=1.0\linewidth]{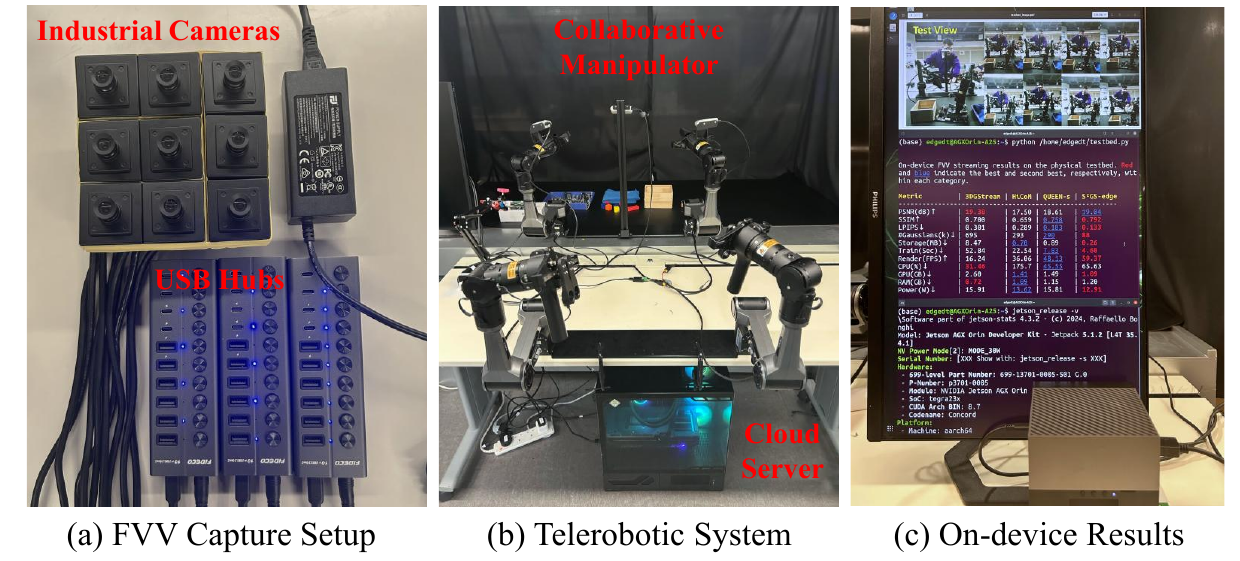}

   \caption{Physical testbed with IoT application of the proposed method.}
   \label{testbed}
\end{figure}

\begin{figure}[t]
  \centering
   \includegraphics[width=1.0\linewidth]{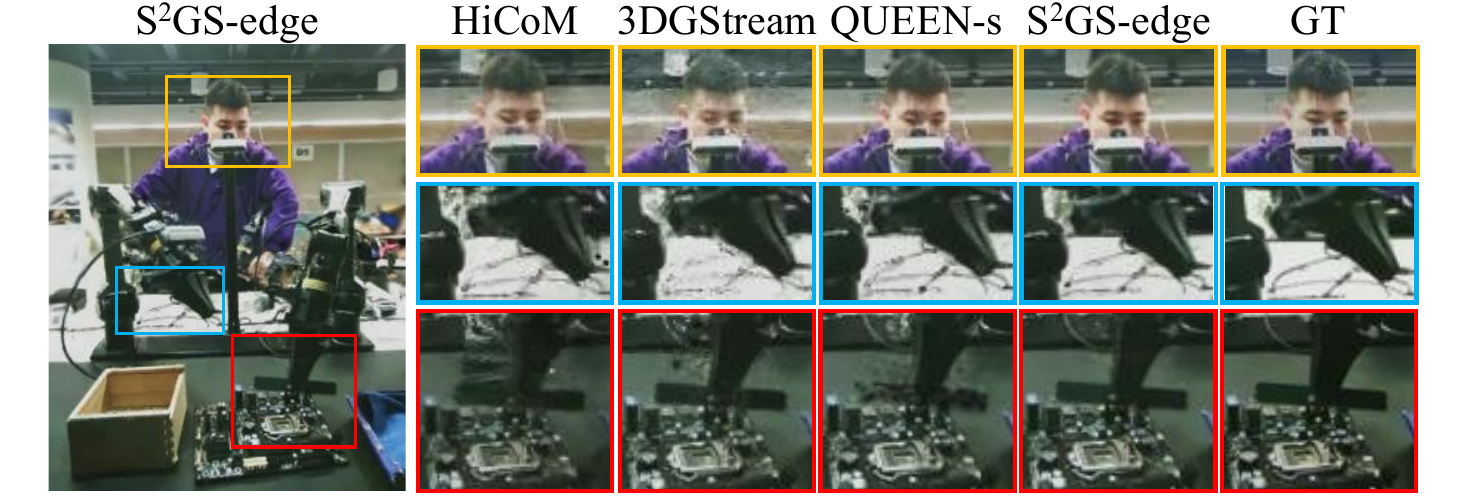}

   \caption{Qualitative comparison with SOTA methods \cite{sun20243dgstream, gao2024hicom, girish2024queen} on the physical testbed. S$^2$GS-edge achieves clearer visual details, demonstrating its practical feasibility for edge-assisted IoT applications.}
   \label{results5}
\end{figure}

The results in Fig.~\ref{testbed}(c) and Fig.~\ref{results5} demonstrate the practical behavior of S$^2$GS-edge. On the physical testbed, S$^2$GS-edge achieves higher visual quality than the baseline methods while maintaining a compact representation with 88k Gaussians and 0.26 MB storage. In terms of efficiency, S$^2$GS-edge reduces the per-frame optimization time to 4.60 s and reaches a rendering throughput of 59.37 FPS on the edge device, with low GPU memory usage and power consumption. These results suggest that the proposed framework can be effectively performed on resource-constrained edge hardware for efficient FVV streaming. Overall, this case study provides proof-of-concept evidence for the practical feasibility of S$^2$GS in a real-world edge-IoT prototype.

\section{Conclusion}
In this paper, we propose S$^2$GS, a structured sparse Gaussian streaming framework for efficient FVV reconstruction. S$^2$GS organizes temporal residuals with a streaming octree and learns sparse residual updates through a differentiable structured gating mechanism. This design reduces storage overhead and per-frame optimization cost while maintaining competitive visual quality. Extensive experiments on benchmark datasets, a consumer GPU, an edge IoT device, and a physical testbed demonstrate the effectiveness and practical feasibility of S$^2$GS.

Nevertheless, S$^2$GS still relies on the quality of first-frame reconstruction and adopts a fixed streaming hierarchy, which limits its adaptability in scenarios involving severe occlusion/disocclusion and large out-of-support motions, as discussed in Appendix \ref{First-Frame Influence} and \ref{Limitations of S$^2$GS}. In addition, its robustness under practical industrial disturbances, such as camera desynchronization \cite{xu2026dynamic} and sparse views \cite{li2025geometry}, remains to be evaluated. Future work will explore hierarchy refresh, keyframe-based refinement \cite{yan2025instant}, and systematic stress testing to enhance robustness for real-world deployment on edge-IoT systems.

\appendices
\section{Supplemental Videos}
\label{appA}
We provide supplemental videos at \href{https://github.com/liyw420/S2GS_SupplementaryMaterial}{this URL} to facilitate assessment of the visual quality of the proposed method. The video provides high-resolution FVV streaming results that are difficult to characterize using quantitative metrics alone. Specifically, the supplemental video includes:

\begin{enumerate}
    \item Streaming FVV reconstruction results of S$^2$GS-full and S$^2$GS-fast on diverse dynamic scene datasets, including N3DV \cite{li2022neural}, MeetRoom \cite{li2022streaming}, and ENeRF Outdoor \cite{lin2022efficient}.
    \item Results of S$^2$GS-edge on the real-world physical testbed.
    \item Representative failure cases of S$^2$GS on more challenging outdoor scenes of the Campus dataset \cite{wang2023masked}.
\end{enumerate}

\section{Implementation Details}
\label{Implementation Details}

For all self-measured baselines, we used the official implementations and the optimization settings recommended by their authors. We held the dataset splits, input resolutions, training and test views, camera parameters, and hardware platform constant across all self-measured methods. For initialization, we followed the official protocol of each method. For methods with the same initialization pipelines, such as QUEEN \cite{girish2024queen} and S$^2$GS, we used the same first-frame COLMAP \cite{schoenberger2016mvs,schoenberger2016sfm} reconstruction and camera parameters. Results taken from original papers are included only as reference values; they are not part of the strictly controlled comparisons under matched hardware and computational budgets.

In addition, to improve transparency and reproducibility, Tab.~\ref{tab:scene_hyperparameters} reports the exact hyperparameter configuration used for each evaluated scene. For the sensitivity analyses of both structural parameters and sparse regularization weights, please refer to Appendix \ref{Streaming Octree Representation} and Appendix \ref{Sparse Regularization}, respectively.

\begin{table}[!t]
\centering
\caption{Per-scene hyperparameter configurations. The physical testbed and Campus dataset are additional evaluations reported in the appendix.}
\label{tab:scene_hyperparameters}

\footnotesize
\setlength{\tabcolsep}{4.0pt}
\renewcommand{\arraystretch}{1.05}

\begin{tabular}{llcccc}
\toprule
Dataset & Scene & \makecell{Root\\Resolution} & \makecell{Log\\Base} & $\lambda_1$ & $\lambda_2$ \\
\midrule
N3DV & Coffee Martini & $2^{12}$ & 2.0 & 0.01 & 0.01 \\
N3DV & Cook Spinach & $2^{9}$ & 2.0 & 0.01 & 0.01 \\
N3DV & Cut Roasted Beef & $2^{9}$ & 2.0 & 0.01 & 0.01 \\
N3DV & Flame Salmon & $2^{12}$ & 2.0 & 0.01 & 0.01 \\
N3DV & Flame Steak & $2^{9}$ & 2.0 & 0.01 & 0.01 \\
N3DV & Sear Steak & $2^{9}$ & 2.0 & 0.01 & 0.01 \\
\midrule
MeetRoom & Discussion & $2^{8}$ & 1.5 & 0.01 & 0.01 \\
MeetRoom & Trimming & $2^{8}$ & 1.5 & 0.01 & 0.01 \\
MeetRoom & Vrheadset & $2^{8}$ & 1.5 & 0.01 & 0.01 \\
\midrule
ENeRF Outdoor & Actor1\_4 & $2^{9}$ & 1.2 & 0.01 & 0.01 \\
ENeRF Outdoor & Actor2\_3 & $2^{9}$ & 1.2 & 0.01 & 0.01 \\
ENeRF Outdoor & Actor5\_6 & $2^{9}$ & 1.2 & 0.01 & 0.01 \\
\midrule
Physical Testbed & Install RAM & $2^{12}$ & 1.5 & 0.01 & 0.01 \\
Physical Testbed & Pick Blocks & $2^{12}$ & 1.5 & 0.01 & 0.01 \\
Physical Testbed & Unplug Device & $2^{12}$ & 1.5 & 0.01 & 0.01 \\
\midrule
Campus & Children & $2^{12}$ & 2.0 & 0.01 & 0.01 \\
Campus & Football & $2^{12}$ & 2.0 & 0.01 & 0.01 \\
Campus & Fountain & $2^{12}$ & 2.0 & 0.01 & 0.01 \\
\bottomrule
\end{tabular}
\end{table}

\section{Definition and Statistical Analysis of Structured Sparsity}
\label{Definition and Statistical Analysis of Structured Sparsity}

In S$^2$GS, \textbf{\textit{sparsity}} is defined by the fraction of candidate Gaussian residual updates that remain active at each streaming frame after gate discretization. A Gaussian primitive is considered active if and only if the discretized gate associated with its structured region is non-zero. We quantify this property using the ratio of active Gaussian primitives to the total number of candidate Gaussians, where a lower active ratio indicates a sparser residual-update pattern. \textbf{\textit{Structured sparsity}} further requires these active residual updates to follow the spatial hierarchy of the scene. Accordingly, active residual updates are organized into spatially correlated regions and controlled by a small number of structured gates, rather than being selected independently at the primitive level.

Let $\widetilde{g}_{t,j}\in(0,1)$ denote the continuous gate produced by Gumbel-Sigmoid sampling for the $j$-th structured region at frame $t$. During the forward pass, it is discretized as:
\begin{equation}
    \bar{g}_{t,j}
    =\operatorname{ML-STE}(\widetilde{g}_{t,j})
    \in
    \{0.0,0.1,\ldots,0.9,1.0\},
\end{equation}

Therefore, the gates applied to residual updates in the forward pass are discrete rather than continuous. Let $N_t$ and $R_t$ denote the numbers of Gaussian primitives and structured gates at frame $t$, respectively, and let $\operatorname{idx}(i)$ map Gaussian $i$ to its associated structured gate. The gated residual update of Gaussian $i$ is calculated as:
\begin{equation}
    \widehat{\Delta\mathbf{r}}_{t,i}
    =
    \bar{g}_{t,\operatorname{idx}(i)}
    \Delta\mathbf{r}_{t,i}.
\end{equation}
Accordingly, a Gaussian is considered active when its associated gate is non-zero. We define the active Gaussian set as:
\begin{equation}
    \mathcal{A}^{\mathrm{G}}_t
    =
    \left\{
        i\in\{1,\ldots,N_t\}
        \,\middle|\,
        \bar{g}_{t,\operatorname{idx}(i)}>0
    \right\}.
\end{equation}
The \textbf{\textit{Active Gaussian Ratio}} is then defined as:
\begin{equation}
    \rho_{\mathrm{gaussian},t}
    =
    \frac{
        \left|\mathcal{A}^{\mathrm{G}}_t\right|
    }{
        N_t
    }.
\end{equation}
This ratio measures primitive-level sparsity: a lower $\rho_{\mathrm{AG},t}$ indicates that fewer Gaussian primitives receive residual updates. Similarly, the active gate set is defined as:
\begin{equation}
    \mathcal{A}^{\mathrm{S}}_t
    =
    \left\{
        j\in\{1,\ldots,R_t\}
        \,\middle|\,
        \bar{g}_{t,j}>0
    \right\}.
\end{equation}
We define the \textbf{\textit{Gate-to-Gaussian Ratio}} as:
\begin{equation}
    \rho_{\mathrm{gate},t}
    =
    \frac{
        \left|\mathcal{A}^{\mathrm{S}}_t\right|
    }{
        \left|\mathcal{A}^{\mathrm{G}}_t\right|
    }.
\end{equation}
This ratio quantifies the extent to which active residual updates are grouped by structured gates. For a given number of active Gaussians, a lower $\rho_{\mathrm{gate},t}$ indicates that fewer structured gates are active and that each active gate covers a larger group of Gaussian primitives. Therefore, a lower ratio reflects a stronger region-wise organization of residual updates.

The two ratios therefore capture complementary aspects of structured sparsity. The Active Gaussian Ratio measures the size of the primitive-level residual-update support, and the Gate-to-Gaussian Ratio measures how compactly this support is organized at the region level. Importantly, whether a residual update is active depends only on whether $\bar{g}_{t,j}>0$. All non-zero gates are treated equally as active when measuring sparsity, while their specific values determine only the scales of the corresponding residual updates. This formulation therefore separates the selection of active residual updates from the control of their update strengths.

In addition, we also conduct \textbf{\textit{statistical analysis}} to provide evidence for structured sparsity in residual updates. Specifically, as shown in the Supplementary Msssssaterial, across the five evaluated datasets, the dataset-level ratio of active Gaussians ranges from 1.99\% to 12.94\%, showing that temporal residual updates involve a small subset of the Gaussian primitives. Moreover, active gates account for only 9.90\% to 23.94\% of the active Gaussians, indicating that these updates are spatially concentrated within a limited number of hierarchical regions. The resolution analysis further shows that this pattern persists from $1/2$ to $1/16$ input resolution, although the degree of sparsity varies with scene dynamics and representation granularity. These results provide direct quantitative support for structured sparsity across diverse evaluation settings.

\begin{table*}[t]
\centering
\caption{Ablation study of the streaming octree in terms of reconstruction quality, parameter efficiency, and training speed. \textcolor{red}{\textbf{Red}} indicates the best result. HRA and ESQ denote hierarchical residual allocation and efficient streaming query, respectively. Experiments are conducted without the HFP module.}
\label{table_octree_ablation}

\fontsize{8.5}{10}\selectfont
\setlength{\tabcolsep}{8.0pt}
\renewcommand{\arraystretch}{1.0}
\begin{tabular}{cc|ccccc|ccccc}
\toprule
\multicolumn{2}{c|}{Streaming Octree} 
& \multicolumn{5}{c|}{N3DV Dataset} 
& \multicolumn{5}{c}{MeetRoom Dataset} \\
\cmidrule(r){1-2}
\cmidrule(r){3-7}
\cmidrule(r){8-12}
\makecell{HRA} 
& \makecell{ESQ}
& \makecell{PSNR \\ (dB)$\uparrow$}
& \makecell{\#Gaussians \\ (k)$\downarrow$}
& \makecell{\#Active \\ Gates$\downarrow$}
& \makecell{Storage \\ (MB)$\downarrow$}
& \makecell{Train \\ (Sec)$\downarrow$}
& \makecell{PSNR \\ (dB)$\uparrow$}
& \makecell{\#Gaussians \\ (k)$\downarrow$}
& \makecell{\#Active \\ Gates$\downarrow$}
& \makecell{Storage \\ (MB)$\downarrow$}
& \makecell{Train \\ (Sec)$\downarrow$}\\
\midrule
- & - 
& 32.25 & 331 & 16k & 1.10 & 12.7 
& 31.89 & 127 & 13k & 0.22 & 2.83 \\
$\checkmark$ & - 
& \textcolor{red}{\textbf{32.74}} & 122 & 2.7k & 0.34 & 4.58 
& 32.40 & 71 & 2.4k & 0.16 & 2.26 \\
$\checkmark$ & $\checkmark$ 
& 32.71 & \textcolor{red}{\textbf{101}} & \textcolor{red}{\textbf{2.4k}} & \textcolor{red}{\textbf{0.29}} & \textcolor{red}{\textbf{4.24}} 
& \textcolor{red}{\textbf{32.44}} & \textcolor{red}{\textbf{64}} & \textcolor{red}{\textbf{2.2k}} & \textcolor{red}{\textbf{0.14}} & \textcolor{red}{\textbf{2.02}} \\
\bottomrule
\end{tabular}
\end{table*}

\begin{table}[t]
\centering
\caption{Effect of octree root resolution on the MeetRoom dataset.}
\label{root_resolution}

\footnotesize
\setlength{\tabcolsep}{4.0pt}
\renewcommand{\arraystretch}{1.0}
\begin{tabular}{l|cccccc}
\toprule
Resolution 
& \multicolumn{1}{c}{\makecell{PSNR \\ (dB)$\uparrow$}} 
& \multicolumn{1}{c}{\makecell{\#Gaussians \\ (k)$\downarrow$}} 
& \multicolumn{1}{c}{\makecell{\#Active \\ Gates$\downarrow$}} 
& \multicolumn{1}{c}{\makecell{Storage \\ (MB)$\downarrow$}} 
& \multicolumn{1}{c}{\makecell{Train \\ (Sec)$\downarrow$}} 
& \multicolumn{1}{c}{\makecell{Render \\ (FPS)$\uparrow$}} \\
\midrule
$2^{7}$  & 32.39 & 35  & 154 & 0.05 & 1.97 & 526 \\
$2^{8}$  & 32.46 & 64  & 401 & 0.08 & 1.96 & 513 \\
$2^{9}$  & 32.59 & 89  & 828 & 0.11 & 2.02 & 505 \\
$2^{10}$ & 32.41 & 102 & 999 & 0.12 & 2.04 & 498 \\
$2^{11}$ & 32.36 & 106 & 974 & 0.12 & 2.00 & 501 \\
\bottomrule
\end{tabular}
\end{table}

\begin{table}[t]
\centering
\caption{Effect of anchor selection on the MeetRoom dataset.}
\label{table_analysis1}

\footnotesize
\setlength{\tabcolsep}{4.0pt}
\renewcommand{\arraystretch}{1.0}
\begin{tabular}{l|cccccc}
\toprule
Log Base
& \multicolumn{1}{c}{\makecell{PSNR \\ (dB)$\uparrow$}} 
& \multicolumn{1}{c}{\makecell{\#Gaussians \\ (k)$\downarrow$}} 
& \multicolumn{1}{c}{\makecell{\#Active \\ Gates$\downarrow$}} 
& \multicolumn{1}{c}{\makecell{Storage \\ (MB)$\downarrow$}} 
& \multicolumn{1}{c}{\makecell{Train \\ (Sec)$\downarrow$}} 
& \multicolumn{1}{c}{\makecell{Render \\ (FPS)$\uparrow$}} \\
\midrule
$1.2$ & 32.23 & 122 & 1.2k & 0.16 & 2.18 & 495 \\
$1.5$ & 32.46 & 64  & 401  & 0.08 & 1.96 & 513 \\
$2.0$ & 32.29 & 48  & 218  & 0.06 & 2.02 & 501 \\
$2.5$ & 32.33 & 36  & 165  & 0.05 & 1.95 & 519 \\
$3.0$ & 32.20 & 28  & 117  & 0.04 & 1.91 & 525 \\
\bottomrule
\end{tabular}
\end{table}

\begin{table*}[t]
\centering
\caption{Ablation study on the effect of progressive training. \textcolor{red}{\textbf{Red}} indicates the best result.}
\label{table_progressive_training}

\fontsize{8.5}{10}\selectfont
\setlength{\tabcolsep}{8.0pt}
\renewcommand{\arraystretch}{1.0}
\begin{tabular}{c|ccccc|ccccc}
\toprule
\multicolumn{1}{c|}{} 
& \multicolumn{5}{c|}{N3DV Dataset} 
& \multicolumn{5}{c}{ENeRF Outdoor Dataset} \\
\cmidrule(r){2-6}
\cmidrule(r){7-11}
\makecell{Configuration \\ \strut}
& \makecell{PSNR \\ (dB)$\uparrow$}
& \makecell{SSIM \\ $\uparrow$}
& \makecell{\#Active \\ Gates$\downarrow$}
& \makecell{Storage \\ (MB)$\downarrow$}
& \makecell{Train \\ (Sec)$\downarrow$}
& \makecell{PSNR \\ (dB)$\uparrow$}
& \makecell{SSIM \\ $\uparrow$}
& \makecell{\#Active \\ Gates$\downarrow$}
& \makecell{Storage \\ (MB)$\downarrow$}
& \makecell{Train \\ (Sec)$\downarrow$}\\
\midrule
S$^2$GS-full (Ours) 
& \textcolor{red}{\textbf{32.76}} 
& \textcolor{red}{\textbf{0.952}} 
& \textcolor{red}{\textbf{679}} 
& \textcolor{red}{\textbf{0.11}} 
& \textcolor{red}{\textbf{4.12}} 
& 25.94 
& \textcolor{red}{\textbf{0.863}} 
& \textcolor{red}{\textbf{1.9k}} 
& \textcolor{red}{\textbf{0.20}} 
& \textcolor{red}{\textbf{2.28}} \\
w/o Prog. Train. 
& 32.68 
& 0.951 
& 755 
& 0.14 
& 4.15 
& \textcolor{red}{\textbf{26.02}} 
& 0.861 
& 2.1k 
& 0.24 
& 2.33 \\
\bottomrule
\end{tabular}
\end{table*}

\section{Streaming Octree Representation}
\label{Streaming Octree Representation}

We conduct additional experiments to evaluate the effects of the streaming octree. In this representation, the octree is built within the global spatial bounds of the first-frame root Gaussian primitives. Candidate anchors are initialized at voxel centroids across the multi-resolution hierarchy \cite{renoctree}. The position of initial Gaussians is calculated as:
\begin{align}
\label{eq_anchor}
{\boldsymbol{\mu}_i = {\boldsymbol{h}_x} + {\boldsymbol{o}_i} \cdot {\boldsymbol{f}_x}},
\end{align}
where $\boldsymbol{h}_x \in \mathbb{R}^{3}$ is the position of the anchor point $x$ associated with Gaussian $i$, $\boldsymbol{f}_x \in \mathbb{R}^{3}$ represents the anchor-dependent scaling factor. The variable $\boldsymbol{\mu}_{i}$ specifies the position of the $i$-th Gaussian primitive, while $\boldsymbol{o}_{i} \in \mathbb{R}^{3}$ denotes its offset vector.

The streaming octree improves S$^2$GS through two complementary mechanisms: hierarchical residual allocation (HRA) and efficient streaming query (ESQ). HRA organizes temporally sparse Gaussian residual updates according to the spatial hierarchy of the scene, so that dynamic changes are modeled as structured regional variations rather than independent primitive-wise updates. This preserves spatial correlations among dynamic Gaussians while reducing redundant residual parameters. ESQ, built upon the fixed octree hierarchy and persistent pre-allocated anchors, further reduces the overhead of residual lookup and avoids repeated online pruning or densification during per-frame optimization. 

The results in Tab.~\ref{table_octree_ablation} validate these two effects. On the N3DV dataset, adding HRA improves PSNR from 32.25 dB to 32.74 dB and reduces the number of Gaussians, active gates, storage, and training time from 331k, 16k, 1.10 MB, and 12.7s to 122k, 2.7k, 0.34 MB, and 4.58s, respectively. Adding ESQ further reduces these values to 101k, 2.4k, 0.29 MB, and 4.24s, with only a negligible change in PSNR. Similar trends are observed on the MeetRoom dataset, where the full streaming octree design achieves the best PSNR while using the fewest Gaussians and active gates, the lowest storage cost, and the shortest training time. These results demonstrate that the octree improves reconstruction quality mainly by preserving structured spatial correlations, improves parameter efficiency through compact hierarchical allocation, and accelerates training by reducing both primitive-wise optimization redundancy and query overhead.

In addition to the main ablation, additional analyses are conducted to examine the design choices, including the root resolution, anchor selection, and progressive training.

\subsubsection{Root Resolution} 

Root resolution controls the granularity of region-level sparse updates, thereby affecting the trade-off between reconstruction quality and efficiency. As shown in Tab.~\ref{root_resolution}, using a coarse root resolution of $2^{7}$ yields the most compact representation, with 35k Gaussians, 154 active gates, 0.05 MB storage, and 526 FPS rendering speed, but its PSNR is lower than those of finer-resolution settings. Increasing the resolution to $2^{9}$ improves PSNR from 32.39 dB to 32.59 dB, indicating that finer root-level regions can better capture local dynamic variations. However, this also increases the number of Gaussians, active gates, and storage costs. Further increasing the resolution to $2^{10}$ or $2^{11}$ does not improve reconstruction quality, while introducing higher model complexity and slightly slower rendering. These results show that overly coarse roots may limit local dynamics, whereas overly fine roots introduce redundant parameters and gates. We therefore adopt $2^{8}$ on the MeetRoom dataset, which achieves a favorable balance between reconstruction quality, parameter efficiency, training speed, and rendering performance.

\subsubsection{Anchor Selection} We study the heuristic resolution-based anchor selection method \cite{renoctree}. It estimates the view-dependent hierarchy level of anchor points as follows:
\begin{equation}
\hat{h}_{xy}
=
\left\lfloor h^{\mathrm{soft}}_{xy} \right\rfloor
=
\left\lfloor
\Gamma\left(
\log\left(\rho_{\max}/{\rho_{xy}}
\right)
\right)
\right\rfloor ,
\label{eq:hierarchical_anchor_level}
\end{equation}
where $\rho_{xy}$ is the observation distance from camera viewpoint $y$ to anchor point $x$, $\rho_{\max}$ is the reference maximum distance of the octree box, $\Gamma(\cdot)$ is a clamping function to constrain the soft hierarchy level to the valid range, and $\lfloor\cdot\rfloor$ denotes the floor operator. The anchor is selected if its predefined level $h_x$ does not exceed the estimated view-dependent level $\hat{h}_{xy}$.

We conduct an ablation study on the logarithmic base for anchor selection on the MeetRoom dataset. As shown in Tab.~\ref{table_analysis1}, the log base controls the trade-off between representation capacity and compactness. A smaller base, such as $1.2$, retains more Gaussians and active gates, increasing model complexity without achieving the best reconstruction quality. In contrast, a larger base, such as $3.0$, yields the most compact representation and highest rendering speed, but reduces PSNR to 32.20 dB, suggesting insufficient capacity for dynamic details. The base $1.5$ achieves the best PSNR while maintaining moderate Gaussians, active gates, storage, and training time. Therefore, we adopt $1.5$ as the default log base on the MeetRoom dataset, which provides a favorable balance between reconstruction quality and streaming efficiency.

\begin{figure}[t]
\centering
\includegraphics[width=1.0\linewidth]{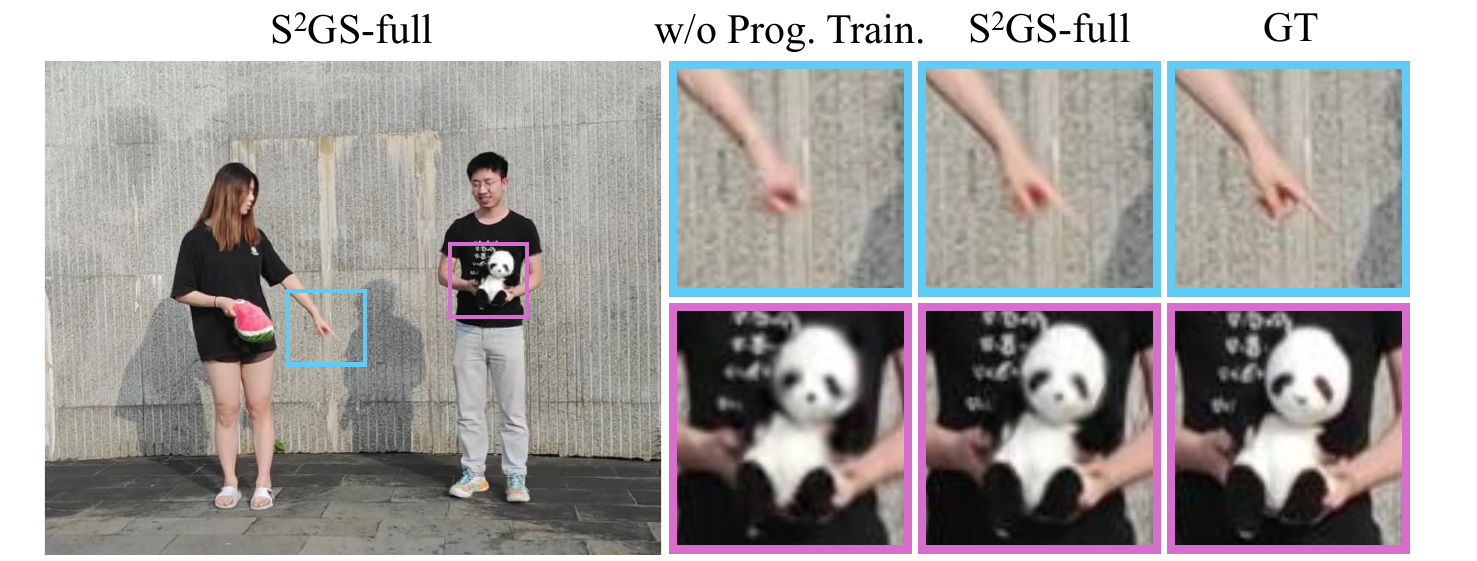}

\caption{Visualizations of rendered images from S$^2$GS-full and the variant without progressive training. Progressive training facilitates the recovery of dynamic objects with fine-grained motions.}
\label{prog_train}
\end{figure}

\subsubsection{Progressive Training} 
We further evaluate the effect of progressive training on reconstruction quality and streaming efficiency. Since the streaming octree represents Gaussian residuals using hierarchical levels, optimizing all levels simultaneously may introduce inherent challenges. In S$^2$GS, we adopt a coarse-to-fine training scheme. Let $H$ denote the total number of residual levels. For each streaming frame, we initially activate residual levels from $\lfloor H/2 \rfloor$, and then progressively introduce finer levels during optimization. Since leaf-level residuals are designed to capture fine-grained local dynamics, we allocate more optimization iterations to finer levels. Specifically, if $N_i$ denotes the number of iterations assigned to level $i$, we set
\begin{equation}
N_{i+1} = \omega N_i, \quad \omega \geq 1,
\label{eq:progressive_iterations}
\end{equation}
where $\omega$ is the growth factor controlling the relative training budget between adjacent levels. Thus, finer residual levels receive a larger optimization budget than coarser levels. 

As shown in Tab.~\ref{table_progressive_training}, progressive training improves the overall quality and efficiency. On the N3DV dataset, it improves PSNR from 32.68/0.951 to 32.76/0.952, while reducing active gates from 755 to 679, storage from 0.14 MB to 0.11 MB, and training time from 4.15s to 4.12s. On the ENeRF Outdoor dataset, although the non-progressive variant obtains a slightly higher PSNR, progressive training achieves better SSIM and lower storage costs. The qualitative results in Fig.~\ref{prog_train} further show that progressive training better preserves fine-grained dynamic details. These results indicate that progressive training improves the visual quality and training efficiency.

\section{First-Frame Influence}
\label{First-Frame Influence}

Streaming FVV reconstruction relies on the first frame for the subsequent residual updates. We therefore analyze the effect of first-frame quality by varying the first-frame optimization budget and adding Gaussian noise during training. Specifically, the noisy first-frame images are generated as:
\begin{equation}
\tilde{\mathbf{I}}_0
=
\operatorname{clip}
\left(
\mathbf{I}_0 + \boldsymbol{\epsilon},
0, 1
\right),
\quad
\epsilon_{u,v,c}
\sim
\mathcal{N}
\left(
0,
\left(\frac{\sigma}{255}\right)^2
\right),
\label{eq:first_frame_noise}
\end{equation}
where $\mathbf{I}_0$ is the normalized first-frame training image, 
$\tilde{\mathbf{I}}_0$ is the corresponding noisy image, 
and $\operatorname{clip}(\cdot,0,1)$ restricts pixel values to the valid range $[0,1]$. The perturbation $\boldsymbol{\epsilon}$ is independently sampled for each pixel $(u,v)$ and color channel $c$, and $\sigma$ denotes the noise strength on an 8-bit intensity scale.

As shown in Tab.~\ref{table_first_frame_budget}, increasing the budget from 50 to 250 epochs improves streaming PSNR from 29.72 dB to 30.07 dB, while the per-frame streaming cost remains stable. Tab.~\ref{table_first_frame_noise} further shows that S$^2$GS remains robust under moderate first-frame degradation. When the noise level increases from $\sigma=0$ to $\sigma=20$, the streaming PSNR decreases only slightly from 33.96 dB to 33.55 dB. Under more severe degradation ($\sigma=30$), the PSNR drops more noticeably to 33.06 dB, reflecting the inherent dependence of residual-based streaming methods on first-frame quality. Overall, S$^2$GS benefits from a reliable first-frame initialization, but does not require it to be error-free. Residual updates, streaming octree, and structure-aware gating jointly help maintain stable streaming performance under first-frame quality degradation.

It should be noted that our design does not remove the inherent dependence of S$^2$GS on first-frame quality, but makes the subsequent streaming process more stable and efficient under practical first-frame initialization. In long-term deployment, severe errors in the initial frame or substantial accumulated scene changes may reduce the effectiveness of subsequent residual updates. A practical solution is to periodically refresh the framework using selected keyframes when reconstruction errors increase, or scene changes become significant \cite{chen2025motion, yan2025instant}. This keyframe refresh strategy is orthogonal to S$^2$GS and can be integrated with the proposed FVV framework to improve robustness in long-term streaming scenarios.

\begin{table}[t]
\centering
\caption{Effect of the first-frame training budget on the Coffee Martini scene. Epochs indicate the number of optimization epochs used for the first frame.}
\label{table_first_frame_budget}

\footnotesize
\setlength{\tabcolsep}{4pt}
\renewcommand{\arraystretch}{1.0}
\begin{tabular}{l|cc|ccccc}
\toprule
\makecell{}
& \multicolumn{2}{c|}{First-Frame Metric}
& \multicolumn{5}{c}{Streaming Performance} \\
\cmidrule(r){2-3}
\cmidrule(r){4-8}
\makecell{Epochs \\ \strut}
& \makecell{PSNR \\ (dB)$\uparrow$}
& \makecell{Train \\ (Sec)$\downarrow$}
& \makecell{PSNR \\ (dB)$\uparrow$}
& \makecell{LPIPS \\ $\downarrow$}
& \makecell{Storage \\ (MB)$\downarrow$}
& \makecell{Train \\ (Sec)$\downarrow$}
& \makecell{Render \\ (FPS)$\uparrow$} \\
\midrule
50 
& 26.92 & 9.15
& 29.72 & 0.148 & 0.16 & 3.64 & 475 \\
100
& 28.18 & 19.75
& 29.80 & 0.139 & 0.16 & 3.65 & 460 \\
150
& 28.93 & 27.68
& 29.83 & 0.135 & 0.16 & 3.70 & 462 \\
200 
& 29.26 & 37.99
& 29.95 & 0.134 & 0.17 & 3.74 & 477 \\
250  
& 28.96 & 46.69
& 30.07 & 0.133 & 0.17 & 3.82 & 469 \\
\bottomrule
\end{tabular}
\end{table}

\begin{table}[t]
\centering
\caption{Effect of Gaussian noise degradation during first-frame training on the Cook Spinach scene.}
\label{table_first_frame_noise}

\footnotesize
\setlength{\tabcolsep}{4pt}
\renewcommand{\arraystretch}{1.0}
\begin{tabular}{l|cc|ccccc}
\toprule
\makecell{}
& \multicolumn{2}{c|}{First-Frame Metric}
& \multicolumn{5}{c}{Streaming Performance} \\
\cmidrule(r){2-3}
\cmidrule(r){4-8}
\makecell{Noise \\ \strut}
& \makecell{PSNR \\ (dB)$\uparrow$}
& \makecell{Train \\ (Sec)$\downarrow$}
& \makecell{PSNR \\ (dB)$\uparrow$}
& \makecell{LPIPS \\ $\downarrow$}
& \makecell{Storage \\ (MB)$\downarrow$}
& \makecell{Train \\ (Sec)$\downarrow$}
& \makecell{Render \\ (FPS)$\uparrow$} \\
\midrule
$\alpha=0$
& 33.54 & 29.91
& 33.96 & 0.134 & 0.07 & 4.20 & 498 \\
$\alpha=10$
& 33.30 & 30.22
& 33.74 & 0.135 & 0.07 & 4.23 & 492 \\
$\alpha=20$
& 30.89 & 29.84
& 33.55 & 0.138 & 0.07 & 4.21 & 489 \\
$\alpha=30$
& 28.93 & 28.86
& 33.06 & 0.144 & 0.07 & 4.26 & 495 \\
\bottomrule
\end{tabular}
\end{table}

\section{Dynamic Confidence Estimator}
\label{DynamicConfidenceEstimator}

We estimate the dynamic confidence of each Gaussian based on the viewspace gradient difference (VGD)~\cite{girish2024queen, chen2025motion}. Specifically, during FVV streaming, we consider two consecutive frames with dynamic changes. Given the rendered image $\mathcal{I}^{r}_{t-1}$ at frame $t-1$, we compute the mean squared error (MSE) between $\mathcal{I}^{r}_{t-1}$ and the ground-truth images $\mathcal{I}^{gt}_{t-1}$ and $\mathcal{I}^{gt}_{t}$:
\begin{align}
\mathcal{E}_{t} =  \operatorname{MSE}(\mathcal{I}^{r-1}_{t}, \mathcal{I}^{gt}_{t}), \:\:\: \mathcal{E}_{t-1} =  \operatorname{MSE}(\mathcal{I}^{r}_{t-1}, \mathcal{I}^{gt}_{t-1}).
\end{align}
Then the VGD $\boldsymbol{m} \in \mathbb{R}^{N \times 2}$ from two consecutive frames is calculated as:
\begin{align}
    \boldsymbol{m} = \frac{1}{V} \sum_{v=0}^{V-1}
    \Biggl[
        \frac{\partial \mathcal{E}^{(v)}_t}{\partial \boldsymbol{\mu}^{(v)}_{t-1}}
        \;-\;
        \frac{\partial \mathcal{E}^{(v)}_{t-1}}{\partial \boldsymbol{\mu}^{(v)}_{t-1}}
    \Biggr]
    \label{eq:gradient_difference}
\end{align}
where $\boldsymbol{\mu}^{(v)}_{t-1} \in {{\mathbb{R}}^{N\times2}}$ denotes the projected 2D positions of the Gaussians from the previous frame under view $v$, and $V$ denotes the number of training views.

We investigate the robustness of the VGD-based dynamic confidence estimator. First, we compare the proposed robust VGD estimator with several alternative confidence estimators, including uniform random confidence, constant confidence assignments, and min-max normalized VGD. As shown in Tab.~\ref{table_vgd_estimator}, Constant-Zero and Min-Max VGD produce overly sparse gate decisions, resulting in only 0 or 1 active gates and substantially degraded reconstruction quality. In contrast, Constant-One activates all gates and increases storage and training cost, while providing no quality advantage. Uniform Random achieves reasonable reconstruction quality but requires more active gates and larger storage than our method. Robust VGD achieves the highest PSNR with only 401 active gates and 0.08 MB of storage, demonstrating a more favorable trade-off among reconstruction quality, sparsity, and efficiency. These results indicate that Robust VGD provides more balanced confidence estimation for sparse residual learning.

Additionally, we evaluate the effectiveness of Robust VGD under two representative perturbation factors, namely controlled illumination changes and injected gradient noise. 

\begin{figure}[t]
\centering
\includegraphics[width=1.0\linewidth]{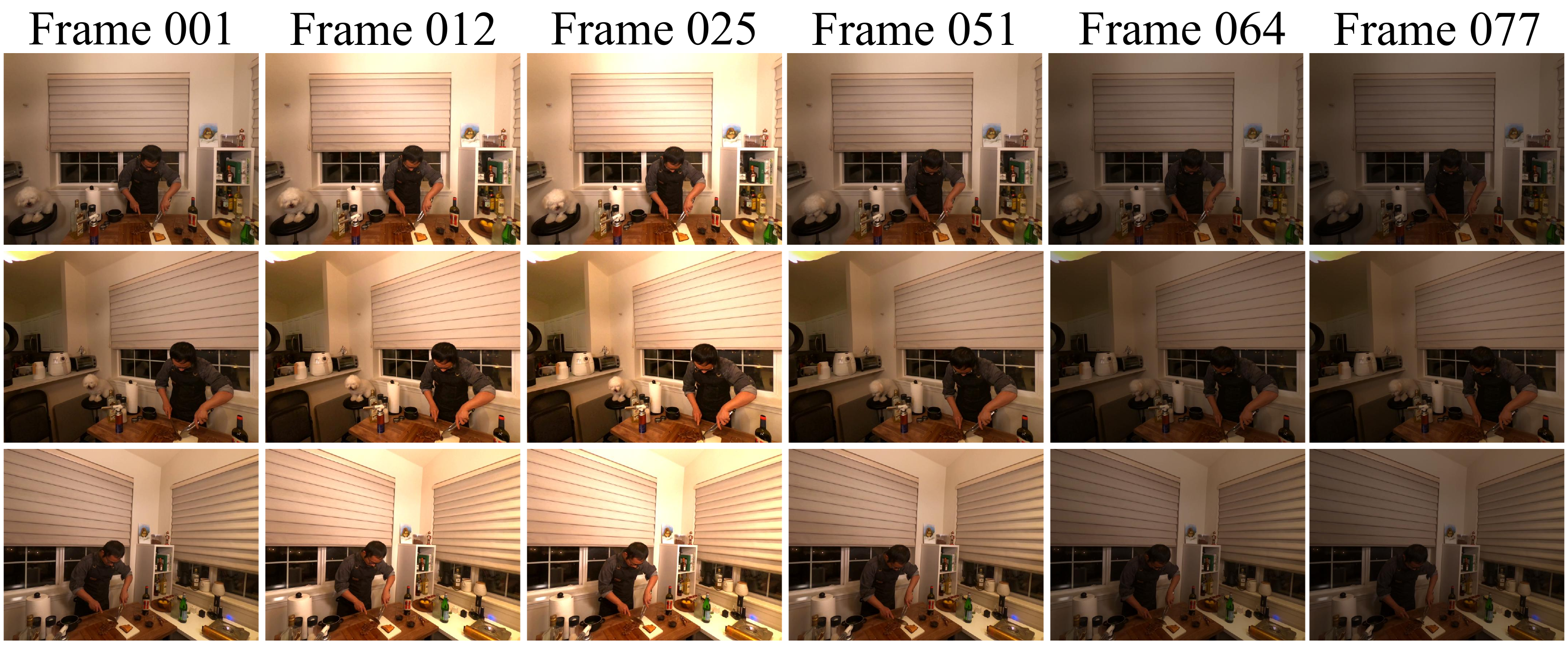}

\caption{Controlled illumination changes on the Cut Roasted Beef scene with a time-varying global gain $g_t \in [0.5, 1.5]$.}
\label{fig:illumination_results}
\end{figure}

\subsubsection{Illumination Changes} We evaluate S$^2$GS under controlled illumination changes by applying a time-varying global gain $g_t$ to the input frames, as illustrated in Fig. \ref{fig:illumination_results}. Results in Tab.~\ref{table_illumination_robustness} show that S$^2$GS maintains stable sparse residual updates across different gain ranges. Compared with the unperturbed setting, the number of active gates changes moderately from 744 to 713/777, while storage remains between 0.07 and 0.08 MB and per-frame training time varies slightly from 4.10 s to 4.21 s. The reconstruction metrics also remain stable under stronger illumination changes. These results indicate that global photometric perturbations do not substantially disrupt VGD-based confidence, suggesting that the proposed robust normalization can maintain stable sparse residual learning under controlled illumination variations.

\subsubsection{Gradient Perturbations} We further evaluate the robustness of S$^2$GS to noisy VGD values by injecting additive Gaussian noise into the VGD estimates. Specifically, the noisy gradient difference is generated as:
\begin{equation}
\tilde{\mathbf{m}}
=
\max\left(
\mathbf{m} + \boldsymbol{\epsilon},
0
\right),
\quad
\boldsymbol{\epsilon}
\sim
\mathcal{N}
\left(
\mathbf{0},
\left(\alpha \cdot \operatorname{Std}(\mathbf{m})\right)^2
\right),
\label{eq:noisy_vgd}
\end{equation}
where $\mathbf{m}$ and $\tilde{\mathbf{m}}$ denote the original and perturbed VGD, respectively. The noise term $\boldsymbol{\epsilon}$ is sampled from a zero-mean Gaussian distribution with standard deviation $\alpha \cdot \operatorname{Std}(\mathbf{m})$, where $\alpha$ controls the relative noise strength.

As shown in Tab.~\ref{table_gradient_robustness}, S$^2$GS remains stable across different noise levels. When the noise strength increases from $\alpha=0$ to $\alpha=0.5$, PSNR changes only slightly from 30.26 dB to 30.13 dB, while SSIM and LPIPS remain nearly unchanged. The number of active gates decreases from 235 to around 180-190, suggesting that noisy confidences lead to more aggressive gate suppression rather than unstable gate decisions. Storage and training time also remain stable. These results suggest that the proposed robust normalization can tolerate gradient perturbations and maintain reliable sparse residual learning under noisy VGD estimates.

\subsubsection{Summary} These results demonstrate that Robust VGD provides a reliable confidence estimator for sparse residual learning, achieving a favorable trade-off among reconstruction quality, storage cost, and training efficiency. The illumination and gradient perturbation analysis further show that median/MAD normalization helps maintain stable sparse residual updates under disturbances. However, VGD is a practical proxy rather than a ground-truth motion estimator. Photometric changes, view inconsistency~\cite{li2025geometry,somraj2024factorized,xu2026dynamic}, severe gradient noise, or weak motions may lead to inaccurate confidence estimation. Thus, our method improves robustness but cannot remove the influence of non-motion factors.

\begin{table}[t]
\centering
\caption{Ablation study of different dynamic confidence estimators on the MeetRoom dataset. Min-Max VGD denotes raw VGD normalized using min-max scaling.}
\label{table_vgd_estimator}

\footnotesize
\setlength{\tabcolsep}{3.0pt}
\renewcommand{\arraystretch}{1.1}
\begin{tabular}{l|cccccc}
\toprule
Confidence Estimator
& \makecell{PSNR \\ (dB)$\uparrow$}
& \makecell{SSIM \\ $\uparrow$}
& \makecell{LPIPS \\ $\downarrow$}
& \makecell{\#Active \\ Gates $\downarrow$}
& \makecell{Storage \\ (MB)$\downarrow$}
& \makecell{Train \\ (Sec)$\downarrow$} \\
\midrule
Uniform Random
& 32.33 & 0.964 & 0.147 & 801 & 0.12 & 2.09 \\
Constant-One
& 32.20 & 0.963 & 0.143 & 12k & 0.63 & 2.48 \\
Constant-Zero
& 25.13 & 0.919 & 0.201 & 0 & 0.04 & 1.91 \\
Min-Max VGD
& 26.03 & 0.923 & 0.197 & 1 & 0.04 & 1.92 \\
Robust VGD (Ours) 
& 32.46 & 0.964 & 0.147 & 401 & 0.08 & 1.96 \\
\bottomrule
\end{tabular}
\end{table}

\begin{table}[t]
\centering
\caption{Robustness analysis under controlled illumination changes on the Cut Roasted Beef scene.}
\label{table_illumination_robustness}

\footnotesize
\setlength{\tabcolsep}{4.0pt}
\renewcommand{\arraystretch}{1.1}
\begin{tabular}{l|cccccc}
\toprule
Illumination Gain
& \makecell{PSNR \\ (dB)$\uparrow$}
& \makecell{SSIM \\ $\uparrow$}
& \makecell{LPIPS \\ $\downarrow$}
& \makecell{\#Active \\ Gates $\downarrow$}
& \makecell{Storage \\ (MB)$\downarrow$}
& \makecell{Train \\ (Sec)$\downarrow$} \\
\midrule
$g_t =1.0$
& 34.12 & 0.958 & 0.133 & 744 & 0.07 & 4.10 \\
$g_t \in [0.8, 1.2]$
& 34.28 & 0.958 & 0.132 & 713 & 0.07 & 4.11 \\
$g_t \in [0.6, 1.4]$
& 34.41 & 0.958 & 0.128 & 753 & 0.08 & 4.16 \\
$g_t \in [0.5, 1.5]$
& 34.98 & 0.958 & 0.123 & 777 & 0.08 & 4.21 \\
\bottomrule
\end{tabular}
\end{table}

\begin{table}[t]
\centering
\caption{Robustness analysis under injected gradient noise on the Flame Salmon scene.}
\label{table_gradient_robustness}

\footnotesize
\setlength{\tabcolsep}{4.0pt}
\renewcommand{\arraystretch}{1.1}
\begin{tabular}{l|cccccc}
\toprule
Injected Noise
& \makecell{PSNR \\ (dB)$\uparrow$}
& \makecell{SSIM \\ $\uparrow$}
& \makecell{LPIPS \\ $\downarrow$}
& \makecell{\#Active \\ Gates $\downarrow$}
& \makecell{Storage \\ (MB)$\downarrow$}
& \makecell{Train \\ (Sec)$\downarrow$} \\
\midrule
$\alpha = 0$
& 30.26 & 0.939 & 0.109 & 235 & 0.18 & 4.17 \\
$\alpha = 0.1$
& 30.18 & 0.939 & 0.109 & 181 & 0.16 & 4.19 \\
$\alpha = 0.2$
& 30.25 & 0.939 & 0.110 & 179 & 0.16 & 4.19 \\
$\alpha = 0.3$
& 30.19 & 0.939 & 0.108 & 182 & 0.16 & 4.18 \\
$\alpha = 0.5$
& 30.13 & 0.938 & 0.109 & 189 & 0.16 & 4.20 \\
\bottomrule
\end{tabular}
\end{table}

\section{Propagation Strategy Analysis}
\label{Propagation Strategy Analysis}

To further validate the design of the proposed hierarchical feature propagation (HFP), we implement a fixed-threshold (FT) pruning baseline for comparison. Given the leaf-level dynamic confidence $p_i$ of Gaussian primitive $i$, the FT-based method first applies a threshold $\tau$ to determine whether the corresponding primitive contains sufficient dynamic evidence:
\begin{equation}
    \hat{p}_i =
    \begin{cases}
        p_i, & p_i \geq \tau, \\
        0,   & p_i < \tau,
    \end{cases}
    \label{eq:ft_leaf_pruning}
\end{equation}
where $\tau$ is the predefined pruning threshold. The thresholded confidences are then propagated to the root-level voxel through the same index mapping $\operatorname{idx}(\cdot)$ used in HFP:
\begin{equation}
    p^{\mathrm{FT}}_j
    =
    \max_{\{i \mid \operatorname{idx}(i)=j\}}
    \hat{p}_i,
    \quad
    i \in \{0,1,\ldots,N-1\},
    \label{eq:ft_root_confidence}
\end{equation}
where $j$ indexes the root-level voxels. If all leaf-level confidences within a local subtree fall below $\tau$, the root-level confidence $p^{\mathrm{FT}}_j$ becomes zero, and the corresponding structured region is unlikely to be activated for residual updates.

Unlike HFP, which preserves the strongest dynamic cue in each local subtree, FT-based propagation applies hard thresholding during optimization. Although this strategy reduces the number of active gates, it discards weak but informative dynamic responses. This effect becomes more pronounced with larger $\tau$, as local motions with moderate confidence values may be irreversibly removed. As a result, FT-based propagation produces overly sparse gate decisions, thereby degrading reconstruction quality in dynamic regions. 

In our experiments, we evaluate several threshold values, i.e., $\tau \in \{0.50, 0.80, 0.90, 0.95\}$, while keeping all other components unchanged. The resulting root-level confidences $p^{\mathrm{FT}}_j$ are fed into the same Gumbel-Sigmoid sampling and multi-level STE pipeline as the proposed HFP method. Therefore, this comparison isolates the effect of the confidence propagation strategy.

\begin{table}[t]
\centering
\caption{Sensitivity analysis of the regularization weights on the ENeRF Outdoor dataset. The default setting is $\lambda_1=0.01$ and $\lambda_2=0.01$.}
\label{table_reg_sensitivity}

\footnotesize
\setlength{\tabcolsep}{5pt}
\renewcommand{\arraystretch}{1.1}

\begin{tabular}{cc|cccccc}
\toprule
$\lambda_1$ 
& $\lambda_2$
& \makecell{PSNR \\ (dB)$\uparrow$}
& \makecell{LPIPS \\ $\downarrow$} 
& \makecell{\#Active \\ Gates $\downarrow$}
& \makecell{Storage \\ (MB)$\downarrow$}
& \makecell{Train \\ (Sec)$\downarrow$}
& \makecell{Render \\ (FPS)$\uparrow$} \\
\midrule
\multicolumn{8}{c}{\textit{Effect of $\lambda_1$ for Gate Decision}} \\
\midrule
0    & 0.01 & 25.60 & 0.122 & 12k  & 0.79 & 2.44 & 533 \\
0.01 & 0.01 & 25.94 & 0.124 & 1.9k & 0.20 & 2.28 & 541 \\
0.02 & 0.01 & 25.84 & 0.127 & 1.2k & 0.15 & 2.23 & 539 \\
0.03 & 0.01 & 25.78 & 0.128 & 952  & 0.14 & 2.23 & 547 \\
\midrule
\multicolumn{8}{c}{\textit{Effect of $\lambda_2$ for Offset Magnitude}} \\
\midrule
0.01 & 0    & 25.90 & 0.124 & 2.2k & 0.22 & 2.33 & 535 \\
0.01 & 0.01 & 25.94 & 0.124 & 1.9k & 0.20 & 2.28 & 541 \\
0.01 & 0.02 & 25.88 & 0.125 & 1.6k & 0.18 & 2.24 & 536 \\
0.01 & 0.03 & 25.81 & 0.126 & 1.5k & 0.18 & 2.24 & 539 \\
\bottomrule
\end{tabular}
\end{table}

\section{Sparse Regularization}
\label{Sparse Regularization}

To encourage sparse and compact residual updates, we introduce a regularization term that combines an approximate $L_0$ penalty on gate decisions with an $L_2$ penalty on non-zero offset updates. The term $\pi_i$ provides a differentiable estimate of the probability that the $i$-th residual update is activated. Specifically, given the dynamic confidence $p_i$,
\begin{align}
    \pi_i = 1 - F_{p_i}(\eta),
\end{align}
where $F_{p_i}(\eta)$ denotes the cumulative probability that the sampled gate is smaller than a small threshold $\eta$. Thus, $\pi_i$ can be interpreted as the probability that the corresponding residual gate is non-zero. A smaller $\pi_i$ indicates that the residual is more likely to be suppressed, whereas a larger $\pi_i$ indicates that the residual is more likely to be active in the update.

This formulation provides a differentiable method for counting active residual updates. Directly minimizing the number of non-zero gates would require an $L_0$ objective, which is non-differentiable. Instead, $\sum_i \pi_i$ approximates the expected number of active residual updates and can be optimized by gradient descent. Therefore, the first term $\pi_i \lambda_1$ encourages unnecessary residual updates to be inactive, inducing sparse Gaussian updates. Meanwhile, the second term $\pi_i \lambda_2 \|\Delta o_i\|_2^2$ penalizes the magnitude of residual offsets without sparsification, preventing unstable or excessively large updates. During optimization, the photometric loss preserves residuals that are necessary for reconstructing dynamic regions, while $\mathcal{L}_{reg}$ suppresses redundant and noisy residual updates. As a result, the model learns structured sparse Gaussian residual updates in an end-to-end differentiable manner.

We further add a sensitivity analysis of the regularization weights. The purpose of this experiment is not to exhaustively tune hyperparameters, but to verify the functions of the two terms in $\mathcal{L}_{\mathrm{reg}}$. As shown in Tab.~\ref{table_reg_sensitivity}, $\lambda_1$ primarily controls gate sparsity: removing it increases the number of active gates to 12k and storage to 0.79 MB, whereas larger values progressively reduce active gates and storage but slightly degrade PSNR. This confirms that $\lambda_1$ directly regulates the trade-off between sparsity and quality. In contrast, $\lambda_2$ constrains the magnitude of non-zero offset updates. Increasing $\lambda_2$ results in more compact updates, but stronger penalties reduce reconstruction quality. The default setting, $\lambda_1=0.01$ and $\lambda_2=0.01$, achieves the best PSNR of 25.94 dB with only 1.9k active gates and 0.20 MB storage, indicating a balanced trade-off among quality, sparsity, and storage efficiency.

\begin{table*}[t]
\centering
\caption{Quantitative comparison on the NVIDIA Jetson AGX Orin in the case study. Energy per frame is computed as average power multiplied by per-frame training time. \textcolor{red}{\textbf{Red}} and \textcolor{blue}{\underline{Blue}} indicate the best and second best, respectively.}
\footnotesize
\setlength{\tabcolsep}{4.5pt}
\renewcommand{\arraystretch}{1.0}
\begin{tabular}{
l|  
l|  
ccccccc|
ccccc  
}
\toprule
\multirow{3}{*}{Scenes}
& \multirow{3}{*}{Methods}
& \multicolumn{7}{c|}{Quality \& Efficiency} 
& \multicolumn{5}{c}{Resource Consumption}\\

& & \multicolumn{1}{c}{\makecell{PSNR \\ (dB)$\uparrow$}} & \multicolumn{1}{c}{\makecell{SSIM \\ $\uparrow$}} & \multicolumn{1}{c}{\makecell{LPIPS \\ $\downarrow$}} & \multicolumn{1}{c}{\makecell{\#Gaussians\\ (k)$\downarrow$}} & \multicolumn{1}{c}{\makecell{Storage \\ (MB)$\downarrow$}} & \multicolumn{1}{c}{\makecell{Train \\ (Sec)$\downarrow$}} & \multicolumn{1}{c|}{\makecell{Render \\ (FPS)$\uparrow$}} & \multicolumn{1}{c}{\makecell{CPU\\ (\%)$\downarrow$}} & \multicolumn{1}{c}{\makecell{GPU MEM \\ (GB) $\downarrow$}} & \multicolumn{1}{c}{\makecell{RAM \\ (GB)$\downarrow$}} & \multicolumn{1}{c}{\makecell{Power \\ (W)$\downarrow$}} & \multicolumn{1}{c}{\makecell{Energy \\ (J)$\downarrow$}} \\
\midrule
\multirow{4}{*}{Install RAM}
& 3DGStream\cite{sun20243dgstream} & \textcolor{red}{\textbf{19.38}} & 0.700 & 0.301 & 695 & 8.47 & 52.84 & 16.24 & \textcolor{red}{\textbf{31.46}} & 2.60 & \textcolor{red}{\textbf{0.72}} & 15.91 & 840.96\\
& HiCoM\cite{gao2024hicom} & 17.50 & 0.659 & 0.289 & 293 & \textcolor{blue}{\underline{0.70}} & 22.54 & 36.06 & 175.7 & \textcolor{blue}{\underline{1.41}} & \textcolor{blue}{\underline{1.09}} & \textcolor{blue}{\underline{13.62}} & 306.88\\
& QUEEN-s\cite{girish2024queen} & 18.61 & \textcolor{blue}{\underline{0.758}} & \textcolor{blue}{\underline{0.183}} & \textcolor{blue}{\underline{290}} & 0.89 & \textcolor{blue}{\underline{7.83}} & \textcolor{blue}{\underline{48.13}} & \textcolor{blue}{\underline{45.55}} & 1.49 & 1.15 & 15.81 & \textcolor{blue}{\underline{123.77}}\\
& S²GS-edge (Ours) & \textcolor{blue}{\underline{19.04}} & \textcolor{red}{\textbf{0.792}} & \textcolor{red}{\textbf{0.133}} & \textcolor{red}{\textbf{88}} & \textcolor{red}{\textbf{0.26}} & \textcolor{red}{\textbf{4.60}} & \textcolor{red}{\textbf{59.37}} & 65.63 & \textcolor{red}{\textbf{1.09}} & 1.20 & \textcolor{red}{\textbf{12.91}} & \textcolor{red}{\textbf{59.38}}\\
\midrule
\multirow{4}{*}{Pick Blocks}
& 3DGStream\cite{sun20243dgstream} & \textcolor{red}{\textbf{17.85}} & 0.697 & 0.324 & 518 & 8.43 & 50.52 & 21.63 & \textcolor{red}{\textbf{35.43}} & 2.08 & \textcolor{red}{\textbf{0.75}} & 15.20 & 767.90 \\
& HiCoM\cite{gao2024hicom} & 16.47 & 0.681 & 0.284 & \textcolor{blue}{\underline{252}} & \textcolor{blue}{\underline{0.78}} & 20.63 & 42.18 & 178.8 & \textcolor{blue}{\underline{1.35}} & \textcolor{blue}{\underline{1.19}} & \textcolor{blue}{\underline{13.50}} & 278.51 \\
& QUEEN-s\cite{girish2024queen} & 17.05 & \textcolor{blue}{\underline{0.756}} & \textcolor{blue}{\underline{0.206}} & 254 & 1.04 & \textcolor{blue}{\underline{7.50}} & \textcolor{blue}{\underline{51.45}} & \textcolor{blue}{\underline{48.45}} & 1.41 & 1.36 & 14.98 & \textcolor{blue}{\underline{112.35}} \\
& S²GS-edge (Ours) & \textcolor{blue}{\underline{17.40}} & \textcolor{red}{\textbf{0.789}} & \textcolor{red}{\textbf{0.157}} & \textcolor{red}{\textbf{72}} & \textcolor{red}{\textbf{0.33}} & \textcolor{red}{\textbf{4.34}} & \textcolor{red}{\textbf{65.66}} & 71.28 & \textcolor{red}{\textbf{1.05}} & 1.22 & \textcolor{red}{\textbf{12.71}} & \textcolor{red}{\textbf{55.16}} \\
\midrule
\multirow{4}{*}{{Unplug Device}}
& 3DGStream\cite{sun20243dgstream} & \textcolor{blue}{\underline{22.11}} & 0.780 & 0.259 & 410 & 8.32 & 45.09 & 25.16 & \textcolor{red}{\textbf{36.69}} & 1.87 & \textcolor{red}{\textbf{0.70}} & 15.06 & 679.06 \\
& HiCoM\cite{gao2024hicom} & 18.98 & 0.710 & 0.295 & \textcolor{blue}{\underline{231}} & \textcolor{blue}{\underline{0.65}} & 18.02 & \textcolor{blue}{\underline{51.29}} & 173.16 & \textcolor{blue}{\underline{1.24}} & \textcolor{blue}{\underline{1.12}} & \textcolor{blue}{\underline{13.38}} & 241.11 \\
& QUEEN-s\cite{girish2024queen} & 21.45 & \textcolor{blue}{\underline{0.802}} & \textcolor{blue}{\underline{0.227}} & 240 & 1.07 & \textcolor{blue}{\underline{8.13}} & 48.93 & \textcolor{blue}{\underline{47.30}} & 1.39 & 1.17 & 15.12 & \textcolor{blue}{\underline{122.93}} \\
& S²GS-edge (Ours) & \textcolor{red}{\textbf{23.47}} & \textcolor{red}{\textbf{0.861}} & \textcolor{red}{\textbf{0.168}} & \textcolor{red}{\textbf{80}} & \textcolor{red}{\textbf{0.42}} & \textcolor{red}{\textbf{4.47}} & \textcolor{red}{\textbf{60.32}} & 66.02 & \textcolor{red}{\textbf{1.08}} & 1.19 & \textcolor{red}{\textbf{12.85}} & \textcolor{red}{\textbf{57.44}}\\
\bottomrule
\end{tabular}
\label{table_case_study}
\end{table*}

\begin{figure}[t]
\centering
\includegraphics[width=0.5\textwidth]{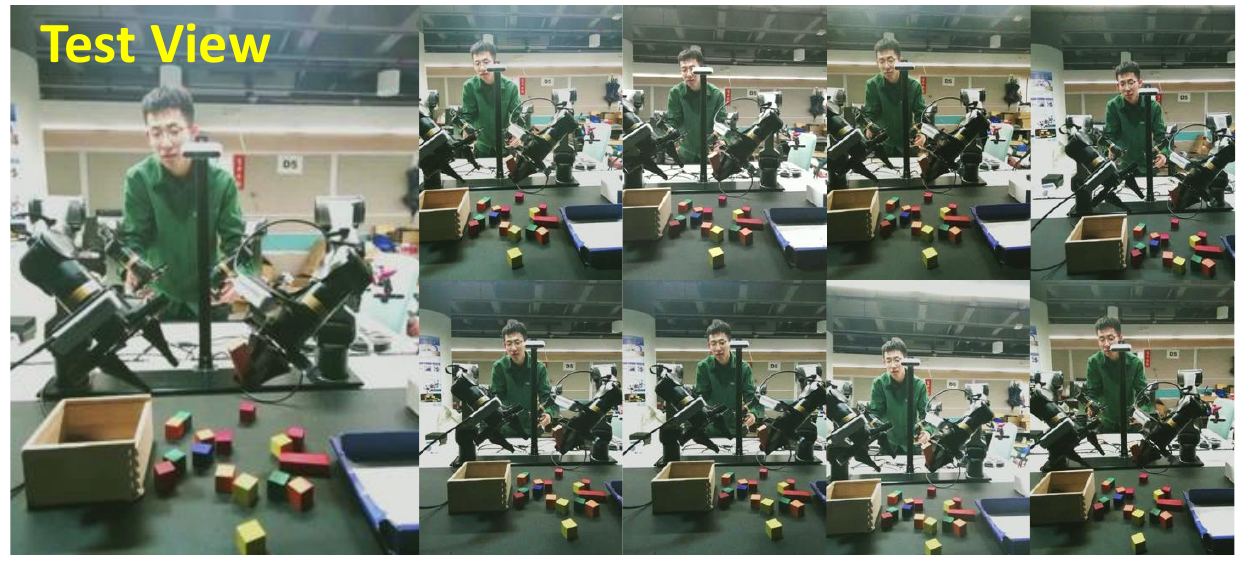}

\caption{Frames from our captured multi-view video (\textit{Pick Blocks} scene). We use 8 camera views for training and the center view for evaluation. The acquisition setup reflects several practical constraints commonly encountered in industrial FVV streaming, including limited viewpoints, low input resolution, and cross-view illumination variations.}
\label{snapshot}
\end{figure}

\begin{figure}[t]
\centering
\includegraphics[width=0.5\textwidth]{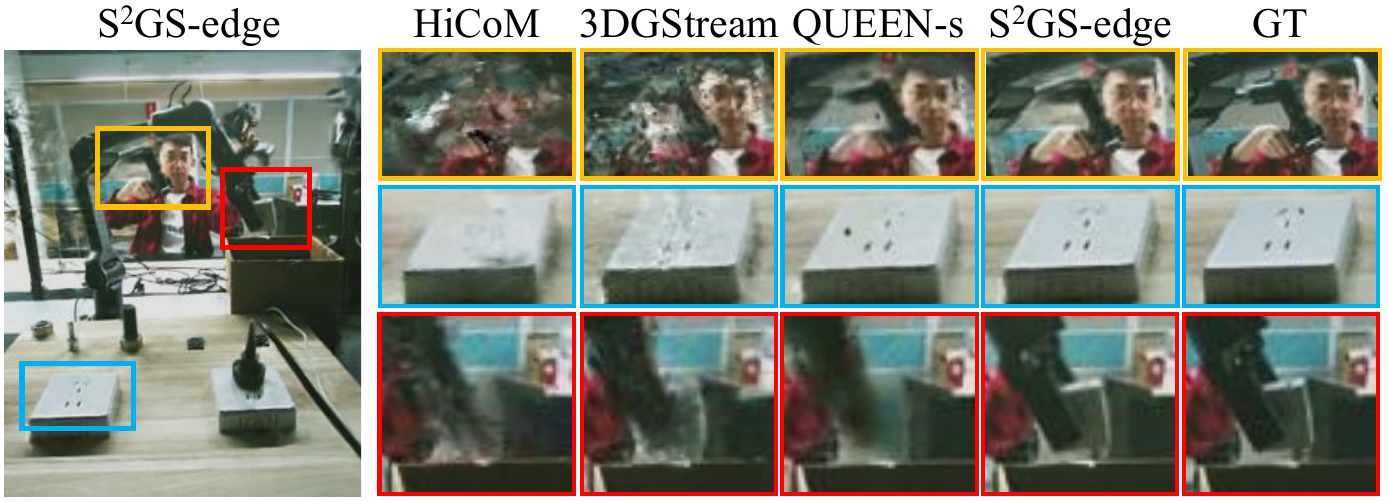}

\caption{Qualitative comparison with SOTA methods \cite{gao2024hicom, sun20243dgstream, girish2024queen} on the \textit{Unplug Device} scene. S2GS-edge achieves clearer results, demonstrating its practical feasibility for edge-assisted IoT applications.}
\label{Unplug_Device}
\end{figure}

\begin{figure*}[t]
\centering
\includegraphics[width=1.0\textwidth]{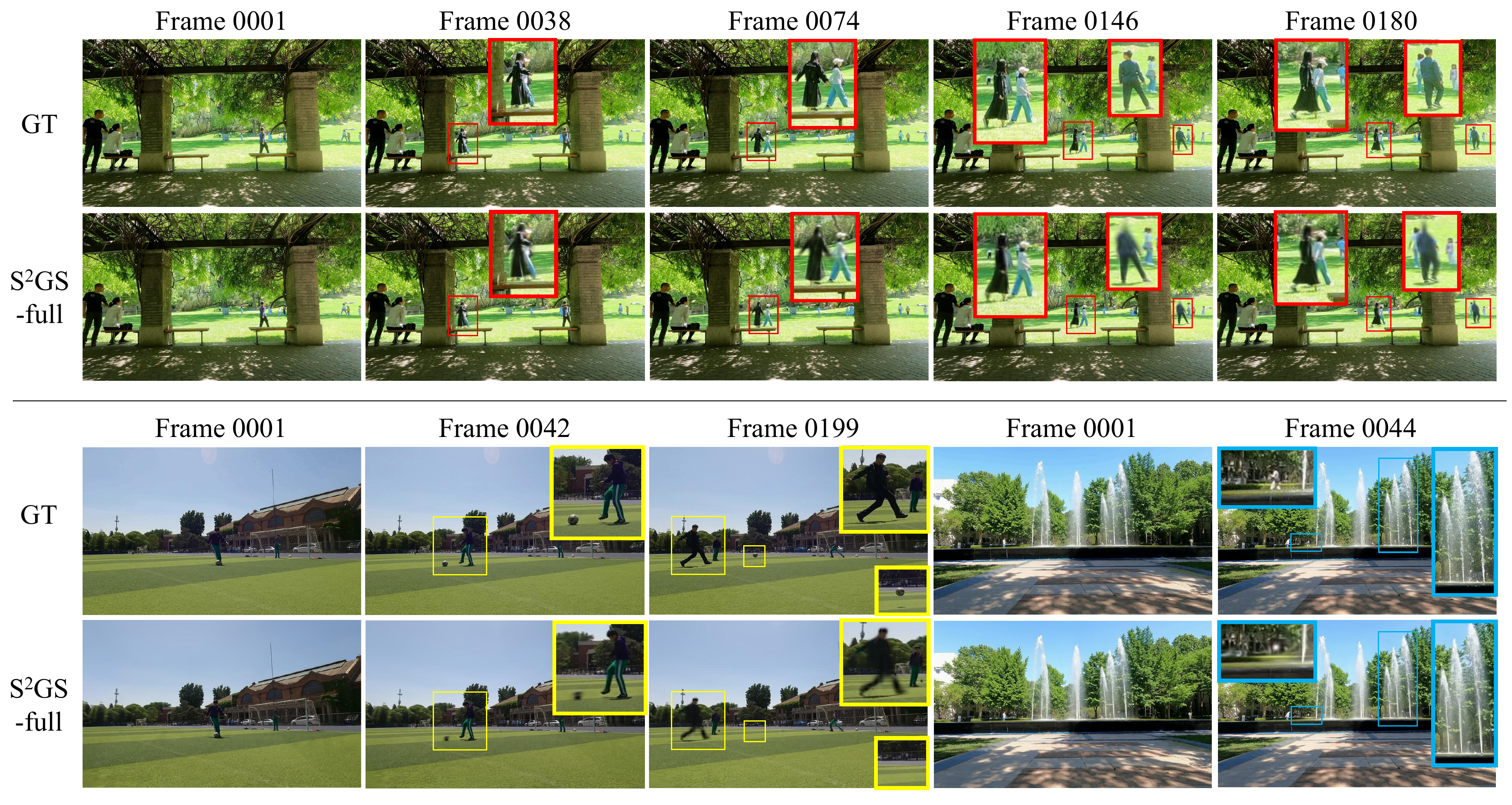}

\caption{Qualitative results on the Campus dataset~\cite{wang2023masked}. 
S$^2$GS may produce blurred or incomplete reconstructions under occlusion/disocclusion, large object motion with out-of-support displacement, and fine-scale stochastic dynamics, as observed in pedestrians emerging from pillars, a fast-moving football with large displacement, and transient splashing fountain droplets.}
\label{campus}
\end{figure*}

\section{Details on Case Study}
\label{Details on Case Study}

We built an FVV streaming system based on a telerobotic platform. The collected dataset contains three telerobotic scenes, namely \textit{Install RAM}, \textit{Pick Blocks}, and \textit{Unplug Device}. A representative snapshot from the \textit{Pick Blocks} sequence is shown in Fig.~\ref{snapshot}. For all scenes, we use 8 camera views for training and the center view for evaluation. As shown in Fig.~\ref{snapshot}, the acquisition setup reflects several practical constraints commonly encountered in industrial FVV streaming, including limited viewpoints, low input resolution, and cross-view illumination changes.

Table~\ref{table_case_study} reports the quantitative results on the NVIDIA Jetson AGX Orin. Across the three telerobotic scenes, S$^2$GS-edge achieves the best SSIM and LPIPS, while maintaining competitive PSNR. It also provides a substantially more compact representation, using only 88k, 72k, and 80k Gaussians with storage footprints of 0.26 MB, 0.33 MB, and 0.42 MB on \textit{Install RAM}, \textit{Pick Blocks}, and \textit{Unplug Device}, respectively. In terms of efficiency, S$^2$GS-edge consistently achieves the shortest per-frame optimization time and the highest rendering throughput, reaching 4.60/4.34/4.47 s and 59.37/65.66/60.32 FPS across the three scenes. 
It further reduces GPU memory usage, power consumption, and per-frame energy cost compared with the baselines. The qualitative comparison in Fig.~\ref{Unplug_Device} shows that S$^2$GS-edge produces clearer details and fewer artifacts, consistent with its quantitative improvements in visual quality. Overall, these results validate S$^2$GS-edge as a proof-of-concept edge deployment for industrial FVV streaming under resource constraints.

\section{Limitations of S$^2$GS}
\label{Limitations of S$^2$GS}

\subsubsection{Fixed Streaming Hierarchy} 
The fixed hierarchy in S$^2$GS improves per-frame efficiency by maintaining persistent anchors and avoiding repeated reconstruction of the hierarchy. This design is effective when scene evolution can be captured by residual updates within the initialized spatial support, but becomes less reliable when later frames require new structural support beyond the fixed hierarchy. To provide concrete evidence, we conduct additional experiments on the challenging Campus dataset~\cite{wang2023masked}. Compared with N3DV, MeetRoom, and ENeRF, Campus contains unbounded outdoor scenes that cover a larger spatial extent and involve more complex dynamics.

As illustrated in Fig.~\ref{campus}, typical failure cases arise with occlusion/disocclusion, large object motion with out-of-support displacement, and fine-scale stochastic dynamics. For example, pedestrians emerging from behind pillars may reveal previously unseen geometry; a fast-moving football may undergo large displacement and move far from its initialized anchors; and splashing fountain droplets contain transient fine-scale structures that rapidly appear and disappear over time. In these cases, S$^2$GS can partially adapt existing Gaussian attributes but cannot create new anchors or reconfigure the spatial hierarchy online, leading to blurred or incomplete renderings. This limitation reflects the trade-off between streaming efficiency and structural adaptivity. Future work will explore anchor insertion or periodic hierarchy refresh to improve robustness to large-scale scene evolution.

\subsubsection{Root-level Gating Strategy} 

The proposed root-level gating strategy determines whether a local hierarchical region receives residual updates, while the associated leaf-level Gaussians retain the capacity to model fine-grained dynamics. However, because all leaf-level Gaussians within the same root region share a common gating decision, a single regional gate may not fully distinguish their individual update requirements when the region contains heterogeneous dynamics, such as multiple independently moving elements or a mixture of static content and localized motion. Activating such a region may trigger residual updates for nearby static Gaussians, thereby reducing the achievable sparsity. Future work will investigate adaptive spatial gating granularity or selective leaf-level gating strategies to provide finer update decisions in highly dynamic and heterogeneous systems \cite{ediger2000spatially, glotzer2000spatially}.

\bibliographystyle{IEEEtran}
\bibliography{reference}

@String{Computing = "Computing" }

@String{Computer = "{IEEE} Computer" }

@String{Springer = "Springer-Verlag" }

@article{li2022streaming,
  title={Streaming radiance fields for 3d video synthesis},
  author={Li, Lingzhi and Shen, Zhen and Wang, Zhongshu and Shen, Li and Tan, Ping},
  journal={Advances in Neural Information Processing Systems},
  volume={35},
  pages={13485--13498},
  year={2022}
}

@inproceedings{furecon,
  title={ReCon-GS: Continuum-Preserved Guassian Streaming for Fast and Compact Reconstruction of Dynamic Scenes},
  author={Fu, Jiaye and Gao, Qiankun and Wen, Chengxiang and Wu, Yanmin and Ma, Siwei and Zhang, Jiaqi and Zhang, Jian},
  booktitle={The Thirty-ninth Annual Conference on Neural Information Processing Systems}
}

@article{mildenhall2021nerf,
  title={Nerf: Representing scenes as neural radiance fields for view synthesis},
  author={Mildenhall, Ben and Srinivasan, Pratul P and Tancik, Matthew and Barron, Jonathan T and others},
  journal={Communications of the ACM},
  volume={65},
  number={1},
  pages={99--106},
  year={2021},
  publisher={ACM New York, NY, USA}
}

@inproceedings{li2022neural,
  title={Neural 3d video synthesis from multi-view video},
  author={Li, Tianye and Slavcheva, Mira and Zollhoefer and others},
  booktitle={Proceedings of the IEEE/CVF conference on computer vision and pattern recognition},
  pages={5521--5531},
  year={2022}
}

@article{song2023nerfplayer,
  title={Nerfplayer: A streamable dynamic scene representation with decomposed neural radiance fields},
  author={Song, Liangchen and Chen, Anpei and Li, Zhong and Chen, Zhang and Chen, Lele and Yuan, Junsong and Xu, Yi and Geiger, Andreas},
  journal={IEEE Transactions on Visualization and Computer Graphics},
  volume={29},
  number={5},
  pages={2732--2742},
  year={2023},
  publisher={IEEE}
}

@article{kerbl20233d,
  title={3D Gaussian splatting for real-time radiance field rendering.},
  author={Kerbl, Bernhard and Kopanas, Georgios and Leimk{\"u}hler, Thomas and Drettakis, George},
  journal={ACM Trans. Graph.},
  volume={42},
  number={4},
  pages={139--1},
  year={2023}
}

@article{girish2024queen,
  title={Queen: Quantized efficient encoding of dynamic gaussians for streaming free-viewpoint videos},
  author={Girish, Sharath and Li, Tianye and Mazumdar, Amrita and Shrivastava, Abhinav and De Mello, Shalini and others},
  journal={Advances in Neural Information Processing Systems},
  volume={37},
  pages={43435--43467},
  year={2024}
}

@inproceedings{sun20243dgstream,
  title={3dgstream: On-the-fly training of 3d gaussians for efficient streaming of photo-realistic free-viewpoint videos},
  author={Sun, Jiakai and Jiao, Han and Li, Guangyuan and Zhang, Zhanjie and Zhao, Lei and Xing, Wei},
  booktitle={Proceedings of the IEEE/CVF Conference on Computer Vision and Pattern Recognition},
  pages={20675--20685},
  year={2024}
}

@article{hu2025varfvv,
  title={Varfvv: View-adaptive real-time interactive free-view video streaming with edge computing},
  author={Hu, Qiang and He, Qihan and Zhong, Houqiang and Lu, Guo and Zhang, Xiaoyun and Zhai, Guangtao and Wang, Yanfeng},
  journal={IEEE Journal on Selected Areas in Communications},
  year={2025},
  publisher={IEEE}
}

@article{gao2024hicom,
  title={Hicom: Hierarchical coherent motion for dynamic streamable scenes with 3d gaussian splatting},
  author={Gao, Qiankun and Meng, Jiarui and Wen, Chengxiang and Chen, Jie and Zhang, Jian},
  journal={Advances in Neural Information Processing Systems},
  volume={37},
  pages={80609--80633},
  year={2024}
}

@article{chen2025motion,
  title={Motion Matters: Compact Gaussian Streaming for Free-Viewpoint Video Reconstruction},
  author={Chen, Jiacong and Mao, Qingyu and Bao, Youneng and Meng, Xiandong and Meng, Fanyang and Wang, Ronggang and Liang, Yongsheng},
  journal={arXiv preprint arXiv:2505.16533},
  year={2025}
}

@inproceedings{yan2025instant,
  title={Instant gaussian stream: Fast and generalizable streaming of dynamic scene reconstruction via gaussian splatting},
  author={Yan, Jinbo and Peng, Rui and Wang, Zhiyan and others},
  booktitle={Proceedings of the Computer Vision and Pattern Recognition Conference},
  pages={16520--16531},
  year={2025}
}

@inproceedings{li2025gifstream,
  title={GIFStream: 4D Gaussian-based Immersive Video with Feature Stream},
  author={Li, Hao and Li, Sicheng and Gao, Xiang and Batuer, Abudouaihati and Yu, Lu and Liao, Yiyi},
  booktitle={Proceedings of the Computer Vision and Pattern Recognition Conference},
  pages={21761--21770},
  year={2025}
}

@inproceedings{hu20254dgc,
  title={4DGC: Rate-Aware 4D Gaussian Compression for Efficient Streamable Free-Viewpoint Video},
  author={Hu, Qiang and Zheng, Zihan and Zhong, Houqiang and Fu, Sihua and Song, Li and Zhang, Xiaoyun and Zhai, Guangtao and Wang, Yanfeng},
  booktitle={Proceedings of the Computer Vision and Pattern Recognition Conference},
  pages={875--885},
  year={2025}
}

@article{renoctree,
  title={Octree-GS: Towards Consistent Real-time Rendering with LOD-Structured 3D Gaussians},
  author={Ren, Kerui and Jiang, Lihan and Lu, Tao and Yu, Mulin and Xu, Linning and Ni, Zhangkai and Dai, Bo},
  journal={IEEE transactions on pattern analysis and machine intelligence}
}

@inproceedings{lu2024scaffold,
  title={Scaffold-gs: Structured 3d gaussians for view-adaptive rendering},
  author={Lu, Tao and Yu, Mulin and Xu, Linning and Xiangli, Yuanbo and Wang, Limin and Lin, Dahua and Dai, Bo},
  booktitle={Proceedings of the IEEE/CVF Conference on Computer Vision and Pattern Recognition},
  pages={20654--20664},
  year={2024}
}

@inproceedings{schoenberger2016sfm,
    author={Sch\"{o}nberger, Johannes Lutz and Frahm, Jan-Michael},
    title={Structure-from-Motion Revisited},
    booktitle={Conference on Computer Vision and Pattern Recognition (CVPR)},
    year={2016},
}

@inproceedings{schoenberger2016mvs,
    author={Sch\"{o}nberger, Johannes Lutz and Zheng, Enliang and Pollefeys, Marc and Frahm, Jan-Michael},
    title={Pixelwise View Selection for Unstructured Multi-View Stereo},
    booktitle={European Conference on Computer Vision (ECCV)},
    year={2016},
}

@inproceedings{maddison2017concrete,
  title={The concrete distribution: A continuous relaxation of discrete random variables},
  author={Maddison, C and Mnih, A and Teh, Y},
  booktitle={Proceedings of the international conference on learning Representations},
  year={2017},
  organization={International Conference on Learning Representations}
}

@article{fang2024maskllm,
  title={Maskllm: Learnable semi-structured sparsity for large language models},
  author={Fang, Gongfan and Yin, Hongxu and Muralidharan, Saurav and Heinrich, Greg and Pool, Jeff and Kautz, Jan and Molchanov, Pavlo and Wang, Xinchao},
  journal={Advances in Neural Information Processing Systems},
  volume={37},
  pages={7736--7758},
  year={2024}
}

@article{bengio2013estimating,
  title={Estimating or propagating gradients through stochastic neurons for conditional computation},
  author={Bengio, Yoshua and L{\'e}onard, Nicholas and Courville, Aaron},
  journal={arXiv preprint arXiv:1308.3432},
  year={2013}
}

@inproceedings{fridovich2023k,
  title={K-planes: Explicit radiance fields in space, time, and appearance},
  author={Fridovich-Keil, Sara and Meanti, Giacomo and Warburg, Frederik Rahb{\ae}k and Recht, Benjamin and Kanazawa, Angjoo},
  booktitle={Proceedings of the IEEE/CVF Conference on Computer Vision and Pattern Recognition},
  pages={12479--12488},
  year={2023}
}

@inproceedings{wu20244d,
  title={4d gaussian splatting for real-time dynamic scene rendering},
  author={Wu, Guanjun and Yi, Taoran and Fang, Jiemin and Xie, Lingxi and Zhang, Xiaopeng and Wei, Wei and Liu, Wenyu and Tian, Qi and Wang, Xinggang},
  booktitle={Proceedings of the IEEE/CVF conference on computer vision and pattern recognition},
  pages={20310--20320},
  year={2024}
}

@article{xu2024grid4d,
  title={Grid4d: 4d decomposed hash encoding for high-fidelity dynamic gaussian splatting},
  author={Xu, Jiawei and Fan, Zexin and Yang, Jian and Xie, Jin},
  journal={Advances in Neural Information Processing Systems},
  volume={37},
  pages={123787--123811},
  year={2024}
}

@inproceedings{li2024spacetime,
  title={Spacetime gaussian feature splatting for real-time dynamic view synthesis},
  author={Li, Zhan and Chen, Zhang and Li, Zhong and Xu, Yi},
  booktitle={Proceedings of the IEEE/CVF Conference on Computer Vision and Pattern Recognition},
  pages={8508--8520},
  year={2024}
}

@article{yang2023real,
  title={Real-time photorealistic dynamic scene representation and rendering with 4d gaussian splatting},
  author={Yang, Zeyu and Yang, Hongye and Pan, Zijie and Zhang, Li},
  journal={arXiv preprint arXiv:2310.10642},
  year={2023}
}

@inproceedings{bae2024per,
  title={Per-gaussian embedding-based deformation for deformable 3d gaussian splatting},
  author={Bae, Jeongmin and Kim, Seoha and Yun, Youngsik and Lee, Hahyun and Bang, Gun and Uh, Youngjung},
  booktitle={European Conference on Computer Vision},
  pages={321--335},
  year={2024},
  organization={Springer}
}

@article{lee2024fully,
  title={Fully explicit dynamic gaussian splatting},
  author={Lee, Junoh and Won, ChangYeon and Jung, Hyunjun and Bae, Inhwan and Jeon, Hae-Gon},
  journal={Advances in Neural Information Processing Systems},
  volume={37},
  pages={5384--5409},
  year={2024}
}

@article{ke2025streamstgs,
  title={StreamSTGS: Streaming Spatial and Temporal Gaussian Grids for Real-Time Free-Viewpoint Video},
  author={Ke, Zhihui and Liu, Yuyang and Zhou, Xiaobo and Qiu, Tie},
  journal={arXiv preprint arXiv:2511.06046},
  year={2025}
}

@inproceedings{wei2025octgpt,
  title={Octgpt: Octree-based multiscale autoregressive models for 3d shape generation},
  author={Wei, Si-Tong and Wang, Rui-Huan and Zhou, Chuan-Zhi and Chen, Baoquan and Wang, Peng-Shuai},
  booktitle={Proceedings of the Special Interest Group on Computer Graphics and Interactive Techniques Conference Conference Papers},
  pages={1--11},
  year={2025}
}

@inproceedings{zheng4dgcpro,
  title={4DGCPro: Efficient Hierarchical 4D Gaussian Compression for Progressive Volumetric Video Streaming},
  author={Zheng, Zihan and Wu, Zhenlong and Zhong, Houqiang and Tian, Yuan and Cao, Ning and Xu, Lan and Yao, Jiangchao and Zhang, Xiaoyun and Hu, Qiang and Zhang, Wenjun},
  booktitle={The Thirty-ninth Annual Conference on Neural Information Processing Systems}
}

@inproceedings{lin2022efficient,
  title={Efficient neural radiance fields for interactive free-viewpoint video},
  author={Lin, Haotong and Peng, Sida and Xu, Zhen and Yan, Yunzhi and Shuai, Qing and Bao, Hujun and Zhou, Xiaowei},
  booktitle={SIGGRAPH Asia 2022 Conference Papers},
  pages={1--9},
  year={2022}
}

@inproceedings{bai2023dynamic,
  title={Dynamic plenoctree for adaptive sampling refinement in explicit nerf},
  author={Bai, Haotian and Lin, Yiqi and Chen, Yize and Wang, Lin},
  booktitle={Proceedings of the IEEE/CVF International Conference on Computer Vision},
  pages={8785--8795},
  year={2023}
}

@misc{jetson_stats,
  author       = {Raffaello Bonghi},
  title        = {jetson-stats},
  howpublished = {\url{https://github.com/rbonghi/jetson_stats}},
  year         = {2023},
  note         = {Accessed: 2026-05-05}
}

@inproceedings{xu2026dynamic,
  title={Dynamic Gaussian Scene Reconstruction from Unsynchronized Videos},
  author={Xu, Zhixin and Zhou, Hengyu and Liu, Yuan and Xue, Wenhan and Pan, Hao and Wang, Wenping and Wang, Bin},
  booktitle={Proceedings of the AAAI Conference on Artificial Intelligence},
  volume={40},
  number={14},
  pages={11469--11477},
  year={2026}
}

@inproceedings{kulhaneklodge,
  title={LODGE: Level-of-Detail Large-Scale Gaussian Splatting with Efficient Rendering},
  author={Kulhanek, Jonas and Rakotosaona, Marie-Julie and Manhardt, Fabian and Tsalicoglou, Christina and Niemeyer, Michael and Sattler, Torsten and Peng, Songyou and Tombari, Federico},
  booktitle={The Thirty-ninth Annual Conference on Neural Information Processing Systems}
}

@article{wei2024fewvv,
  title={Fewvv: Few-shot adaptive bitrate volumetric video streaming with prompted online adaptation},
  author={Wei, Lai and Liu, Yanting and Wang, Fangxin and Wang, Dan},
  journal={IEEE internet of things journal},
  volume={11},
  number={19},
  pages={32055--32066},
  year={2024},
  publisher={IEEE}
}

@article{wei2023vsas,
  title={VSAS: Decision transformer-based on-demand volumetric video streaming with passive frame dropping},
  author={Wei, Lai and Liu, Yanting and Wang, Fangxin and Zhang, Dayou and Wang, Dan},
  journal={IEEE Internet of Things Journal},
  volume={11},
  number={8},
  pages={13752--13767},
  year={2023},
  publisher={IEEE}
}

@article{fu2026generative,
  title={Generative ai-driven digital twin in the manufacturing internet of things: A comprehensive survey},
  author={Fu, Xiuwen and Qin, Meng and Pace, Pasquale and Savaglio, Claudio and Li, Wenfeng and Fortino, Giancarlo},
  journal={IEEE Internet of Things Journal},
  year={2026},
  publisher={IEEE}
}

@article{huang2024enhancing,
  title={Enhancing telecooperation through haptic twin for internet of robotic things: Implementation and challenges},
  author={Huang, Meipeng and Feng, Runhui and Zou, Longhao and Li, Rui and Xie, Jin},
  journal={IEEE Internet of Things Journal},
  volume={11},
  number={20},
  pages={32440--32453},
  year={2024},
  publisher={IEEE}
}

@article{li2022internet,
  title={When internet of things meets metaverse: Convergence of physical and cyber worlds},
  author={Li, Kai and Cui, Yingping and Li, Weicai and Lv, Tiejun and Yuan, Xin and Li, Shenghong and Ni, Wei and Simsek, Meryem and Dressler, Falko},
  journal={IEEE Internet of Things Journal},
  volume={10},
  number={5},
  pages={4148--4173},
  year={2022},
  publisher={IEEE}
}

@article{zhou2026gausshead,
  title={GaussHead: Real-Time 3D Head Avatar Driving for Mobile with Compact Gaussian Representation},
  author={Zhou, Xiaoshi and Yang, Tao and Zhang, Guodong and Song, Baogang and Cao, Chengwei and Bao, Hongwei and Liu, Jialu and Zhang, Yidan and Li, Jing},
  journal={IEEE Internet of Things Journal},
  year={2026},
  publisher={IEEE}
}

@article{tu2026civs,
  title={CIVS: Communication-aware Industrial Video Surveillance with Edge-End Collaboration},
  author={Tu, Jingzheng and Lu, Ying and Yan, Yan and Chen, Cailian and Guan, Xinping},
  journal={IEEE Internet of Things Journal},
  year={2026},
  publisher={IEEE}
}

@article{liao2025graph,
  title={Graph-convolutional-network-enabled task offloading for industrial image recognition in digital twin edge networks},
  author={Liao, Lingling and Tao, Ming and Dong, Ani and Xie, Renping and Zhang, Yin},
  journal={IEEE Internet of Things Journal},
  volume={12},
  number={15},
  pages={29176--29188},
  year={2025},
  publisher={IEEE}
}

@article{wang2023masked,
  title={Masked space-time hash encoding for efficient dynamic scene reconstruction},
  author={Wang, Feng and Chen, Zilong and Wang, Guokang and Song, Yafei and Liu, Huaping},
  journal={Advances in neural information processing systems},
  volume={36},
  pages={70497--70510},
  year={2023}
}

@article{ediger2000spatially,
  title={Spatially heterogeneous dynamics in supercooled liquids},
  author={Ediger, Mark D},
  journal={Annual review of physical chemistry},
  volume={51},
  number={1},
  pages={99--128},
  year={2000},
  publisher={Annual Reviews 4139 El Camino Way, PO Box 10139, Palo Alto, CA 94303-0139, USA}
}

@article{glotzer2000spatially,
  title={Spatially heterogeneous dynamics in liquids: insights from simulation},
  author={Glotzer, Sharon C},
  journal={Journal of Non-Crystalline Solids},
  volume={274},
  number={1-3},
  pages={342--355},
  year={2000},
  publisher={Elsevier}
}

@inproceedings{li2025geometry,
  title={Geometry-Consistent 4D Gaussian Splatting for Sparse-Input Dynamic View Synthesis},
  author={Li, Yiwei and Cao, Jiannong and Ruan, Penghui and Saxena, Divya and Zhu, Songye and Cao, Yinfeng},
  booktitle={2025 IEEE Annual Congress on Artificial Intelligence of Things (AIoT)},
  pages={17--24},
  year={2025},
  organization={IEEE}
}

@inproceedings{somraj2024factorized,
  title={Factorized motion fields for fast sparse input dynamic view synthesis},
  author={Somraj, Nagabhushan and Choudhary, Kapil and Mupparaju, Sai Harsha and Soundararajan, Rajiv},
  booktitle={ACM SIGGRAPH 2024 Conference Papers},
  pages={1--12},
  year={2024}
}

\vfill

\end{document}